\documentclass[10pt,twocolumn,letterpaper]{article}

\usepackage{amsmath,amsfonts,bm}

\newcommand{\etal}{\textit{et al.}}

\def\eqref#1{equation~\ref{#1}}

\def\1{\bm{1}}

\DeclareMathAlphabet{\mathsfit}{\encodingdefault}{\sfdefault}{m}{sl}
\SetMathAlphabet{\mathsfit}{bold}{\encodingdefault}{\sfdefault}{bx}{n}

\def\gF{{\mathcal{F}}}

\def\gO{{\mathcal{O}}}
\def\gP{{\mathcal{P}}}

\renewcommand{\vec}[1]{\boldsymbol{#1}}
\newcommand{\mat}[1]{\mathbf{#1}}
\newcommand{\set}[1]{\mathcal{#1}}

\newcommand{\loss}{\mathcal{L}}
\newcommand{\lossweight}{\lambda}

\newcommand{\geo}{\mathcal{S}}

\newcommand{\point}{\vec{x}}

\newcommand{\normal}{\vec{n}}

\newcommand{\dist}{d}

\newcommand*{\eg}{e.g.\@\xspace}
\newcommand*{\ie}{i.e.\@\xspace}
\usepackage[dvipsnames]{xcolor}
\definecolor{magenta(process)}{rgb}{1.0, 0.0, 0.9}
\newcommand{\heading}[1]{\noindent\textbf{#1.}}

\newcommand{\NICKNAME}{{E3DGE}} 
\newcommand{\nickname}{\NICKNAME} 

\newcommand{\rendererflag}{\mathbf{0}}
\newcommand{\decoderflag}{\mathbf{1}}

\newcommand{\generator}{G}
\newcommand{\rendererG}{\generator_{\rendererflag}}
\newcommand{\decoderG}{\generator_{\decoderflag}}

\newcommand{\encoder}{E}

\newcommand{\encoderGlobal}{\encoder_{\rendererflag}}
\newcommand{\encoderLocal}{\encoder_{\decoderflag}}
\newcommand{\encoderADA}{\encoder_\text{ADA}} %

\newcommand{\zspace}{\mathcal{Z}}
\newcommand{\wspace}{\mathcal{W}}

\newcommand{\surfaceset}{\set{P}_{\gO}}
\newcommand{\additionalset}{\set{P}_{\gF}}

\newcommand{\view}{\bm{v}}

\newcommand{\campose}{\bm\xi}

\newcommand{\setsize}[1]{|{#1}|}

\newcommand{\datasetsample}{\mathcal{X}}

\newcommand{\image}{\mat{I}}

\newcommand{\residual}{\Delta}

\newcommand{\ww}{{\bf w}}

\newcommand{\wcode}{\ww}

\newcommand{\RN}[1]{%
  \textup{\uppercase\expandafter{\romannumeral#1}}%
}

\usepackage{cvpr}              

\usepackage{graphicx}
\usepackage{amsmath}
\usepackage{amssymb}
\usepackage{booktabs}
\usepackage{float}
\usepackage{enumitem}
\usepackage{multirow}

\usepackage[pagebackref,breaklinks,colorlinks]{hyperref}

\usepackage[capitalize]{cleveref}
\crefname{section}{Sec.}{Secs.}
\Crefname{section}{Section}{Sections}
\Crefname{table}{Table}{Tables} 
\crefname{table}{Tab.}{Tabs.}

\def\cvprPaperID{1321} 
\def\confName{CVPR}
\def\confYear{2023}

\begin{document}

\title{Self-Supervised Geometry-Aware Encoder for Style-Based 3D GAN Inversion}

\author{Yushi Lan$^{1}$ \hspace{5mm} Xuyi Meng$^{1}$ \hspace{5mm} Shuai Yang$^{1}$ \hspace{5mm} Chen Change Loy$^{1}$ \hspace{5mm} Bo Dai$^{2}$ \  \\
$^{1}$S-Lab, Nanyang Technological University, Singapore \hspace{8mm}   
$^{2}$Shanghai AI Laboratory \hspace{4mm} 
}

\maketitle

%
\begin{abstract}
StyleGAN has achieved great progress in 2D face reconstruction and semantic editing via image inversion and latent editing. While studies over extending 2D StyleGAN to 3D faces have emerged, a corresponding generic 3D GAN inversion framework is still missing, limiting the applications of 3D face reconstruction and semantic editing.
In this paper, we study the challenging problem of 3D GAN inversion where a latent code is predicted given a single face image to faithfully recover its 3D shapes and detailed textures.
The problem is ill-posed: innumerable compositions of shape and texture could be rendered to the current image. 
Furthermore, with the limited capacity of a global latent code,
2D inversion methods cannot preserve faithful shape and texture at the same time when applied to 3D models.
To solve this problem,  we devise an effective self-training scheme to constrain the learning of inversion. The learning is done efficiently without any real-world 2D-3D training pairs but proxy samples generated from a 3D GAN. 
In addition, apart from a global latent code that captures the coarse shape and texture information,
we augment the generation network with a local branch, where pixel-aligned features are added to faithfully reconstruct face details.
We further consider a new pipeline to perform 3D view-consistent editing.
Extensive experiments show that our method outperforms state-of-the-art inversion methods in both 
shape and texture reconstruction quality.
Code and data will be released.
\end{abstract}
%

\section{Introduction}
\label{sec:intro}

The main goal of this work is to devise an effective approach for encoder-based 3D Generative Adversarial Network (GAN) inversion.
In particular, we focus on the reconstruction of 3D face, requiring just a single 2D face image as the input. 
In the inversion process, we wish to map a given image to the latent space and obtain an editable latent code with an encoder. The latent code will be further fed to a generator to reconstruct the corresponding 3D shape with high-quality shape and texture.
Further to the learning of an inversion encoder, we also wish to develop an approach to synthesize 3D view-consistent editing results, \eg, changing a neutral expression to smiling, by altering the estimated latent code.

GAN inversion~\cite{xia2022gan} has been extensively studied for 2D images but remains underexplored in the 3D world.
Inversion can be achieved via optimization~\cite{abdal2019image2stylegan,abdal2020image2stylegan++,roich2021pivotal}, which typically provides a precise image-to-latent mapping but can be time-consuming, or encoder-based techniques~\cite{richardson2020encoding,wang2021HFGI,tov2021designing}, which explicitly learn an encoding network that maps an image into the latent space.  Encoder-based techniques enjoy faster inversion, but the mapping is typically inferior to optimization.
In this study, we extend the notion of encoder-based inversion from 2D images to 3D shapes. 

Adding the additional dimension makes inversion more challenging beyond the goal of reconstructing an editable shape with detail preservation.
In particular, 
\textbf{1)} Recovering 3D shapes from 2D images is an ill-posed problem,
where innumerable compositions of shape and texture could generate identical rendering results.
3D supervisions are crucial to alleviate the ambiguity of shape inversion from images.
Though high-quality 2D datasets are easily accessible, 
owing to the expensive cost of scans there is currently a lack of large-scale labeled 3D datasets.
\textbf{2)} The global latent code, due to its compact and low-dimensional nature, only captures the coarse shape and texture information. Without high-frequency spatial details, we cannot generate high-fidelity outputs. 
\textbf{3)} Compared with 2D inversion methods where the editing view mostly aligns with the \emph{input view},
in 3D editing we expect the editing results to perform well over the \emph{novel views} with large pose variations.
%
Therefore, 3D GAN inversion is non-trivial task and could not be achieved by directly applying existing approaches.

%
To this end,
we propose a novel \textbf{E}ncoder-based \textbf{3D} \textbf{G}AN inv\textbf{E}rsion framework, \NICKNAME{}, which addresses the aforementioned three challenges.
Our framework has three novel components with a delicate model design. 
Specifically: 

\noindent\textbf{Learning Inversion with Self-supervised Learning} - The first component focuses on the training of the inversion encoder. To address the shape collapse of single-view 3D reconstruction without external 3D datasets,
we retrofit the generator of a 3D GAN model to provide us with diverse pseudo training samples, which can then be used to train our inversion encoder in a self-supervised manner.
Specifically, we generate 3D shapes from the latent space $\wspace$ of a 3D GAN, and then render diverse 2D views from each 3D shape given different camera poses. In this way, we can generate many pseudo 2D-3D pairs together with the corresponding latent codes.
Since the pseudo pairs are generated from a smooth latent space that learns to approximate a natural shape manifold, they serve as effective surrogate data to train the encoder, avoiding potential shape collapse.

\noindent\textbf{Local Features for High-Fidelity Inversion} - The second component learns to reconstruct accurate texture details.
Our novelty here is to leverage local features to enhance the representation capacity, beyond just the global latent code generated by the inversion encoder.
Specifically, in addition to inferring an editable global latent code to represent the overall shape of the face,
we further devise an hour-glass model to extract local features over the residuals details that the global latent code fails to capture.
The local features, with proper projection to the 3D space, serve as conditions to modulate the 2D image rendering.
Through this effective learning scheme, we marry the benefits of both global and local priors and achieve high-fidelity reconstruction.

\noindent\textbf{Synthesizing View-consistent Edited Output} - The third component addresses the problem of novel view synthesis, a problem unique to 3D shape editing. 
Specifically, though we achieve high-fidelity reconstruction through aforementioned designs, the local residual features may not fully align with the scene when being semantically edited. Moreover, the occlusion issue further degrades the fusion performance when rendering from novel views with large pose variations.
To this end, 
we propose a 2D-3D hybrid alignment module for high-quality editing. Specifically, 
a 2D alignment module and a 3D projection scheme are introduced to jointly align the local features with edited images and inpaint occluded local features in novel view synthesis.

Extensive experiments show that our method achieves 3D GAN inversion with plausible shapes and
high-fidelity image reconstruction without affecting editability.
Owing to the self-supervised training strategy with delicate global-local design,
our approach performs well on real-world 2D and 3D benchmarks without resorting to any real-world 3D dataset for training.
%
To summarize, our main contributions are as follows:
\begin{itemize}[itemsep=1.5pt,topsep=2pt,parsep=1.5pt]
\item We propose an early attempt at learning an encoder-based 3D GAN inversion framework for high-quality shape and texture inversion. We show that, with careful design, samples synthesized by a GAN could serve as proxy data for self-supervised training in inversion. 
\item We present an effective framework that uses local features to complement the global latent code for high-fidelity inversion.
\item We propose an effective approach to synthesize view-consistent output through a 2D-3D hybrid alignment module. 
\end{itemize}


\section{Related Work}
\label{sec:related-works}
\heading{3D-aware Image Synthesis}
Generative Adversarial Network~\cite{Goodfellow2014GenerativeAN} has shown promising results in generating photorealistic images~\cite{karras2019style,Brock2019LargeSG,karras_analyzing_2020}
and inspired researchers to put efforts on 3D aware generation~\cite{NguyenPhuoc2019HoloGANUL,platogan,pan_2d_2020}.
However, these methods use explicit shape representations, \ie, voxels~\cite{NguyenPhuoc2019HoloGANUL,platogan} and meshes~\cite{pan_2d_2020} as the intermediate shape models,
which lacks photorealism and is memory-inefficient.
Motivated by the recent success of neural rendering~\cite{park_deepsdf_2019,Mescheder2019OccupancyNetwork,mildenhall2020nerf},
researchers shift to implicit function along with the volume rendering process as the incorporated 3D inductive bias.
Among them, NeRF~\cite{mildenhall2020nerf} proposed an implicit 3D representation for novel view synthesis which defines a scene as $\{\bm{c},\sigma\} = F_{\Phi}(\point, \view)$, 
where $\point$ is the query point, $\view$ is the viewing direction from camera origin to $\point$, $\bm{c}$ is the emitted radiance (RGB value), 
$\sigma$ is the volume density. 
Researchers further extend NeRF to generation task~\cite{Chan2021piGANPI,Schwarz2020NEURIPS} and show impressive view-consistency on the synthesized results.
To increase the generation resolution, recent works~\cite{xu20213dawareis,Chan2021EG3D,eva3d} resort to voxel-based representations or adopting a hybrid design~\cite{GIRAFFE,orel2021stylesdf,Chan2021EG3D,gu2021stylenerf}. By lifting the intermediate low-resolution 2D features to high resolution with a 2D super-resolution decoder,
the hybrid design achieves view-consistent synthesis at high resolution, \eg, $1024^2$.
Beyond synthesizing realistic and diverse images,
previous works~\cite{Besnier2020ThisDD,Pan2022ExploitingDG,jahanian2019steerability,Jahanian2022GenerativeMA,yang2022vtoonify,zhang21datasetgan} have shown that pretrained generators of GAN can be viewed as a compressed and organized training dataset.
Through careful design in the sampling strategy~\cite{Jahanian2022GenerativeMA}, loss functions~\cite{Pan2022ExploitingDG} and generation process~\cite{zhang21datasetgan},
off-the-shelf image generators could facilitate a series of downstream visual applications.

\heading{2D GAN Inversion}
To leverage the strong priors encoded in GANs,
GAN inversion techniques on 2D GANs are well developed.
Optimization-based methods~\cite{abdal2019image2stylegan,abdal2020image2stylegan++} could achieve photorealistic reconstruction at the cost of slow inference and lack of editability.
Encoder-based methods~\cite{richardson2020encoding,wang2021HFGI,tov2021designing,chan2022glean,zhu2020domain} have been developed
to speed up the inversion and show better properties in editing through specific model design~\cite{richardson2020encoding,wang2021HFGI} and training strategies~\cite{tov2021designing}.
pSp \cite{richardson2020encoding} proposed an encoder architecture designed for human faces,
serving as the backbone for many approaches.
e4e \cite{tov2021designing} analyzed the trade-offs between editability and fidelity.
However, they~\cite{richardson2020encoding,tov2021designing,zhu2020domain,abdal2019image2stylegan,abdal2020image2stylegan++} all adopt global latent code alone for GAN inversion task, 
thus failing to recover high-fidelity details.
Recently, 
HFGI~\cite{wang2021HFGI} introduce an extra spatial consultation map to mitigate this issue,
though still designed to restore 2D textures without considering 3D shape modeling.
In this work, we propose a delicate design that exploits local features to recover texture details and achieves view-consistent synthesis.

\section{Preliminaries}
\label{sec:preliminary}

\begin{figure}[t]
    \centering 
    \includegraphics[width=\linewidth]{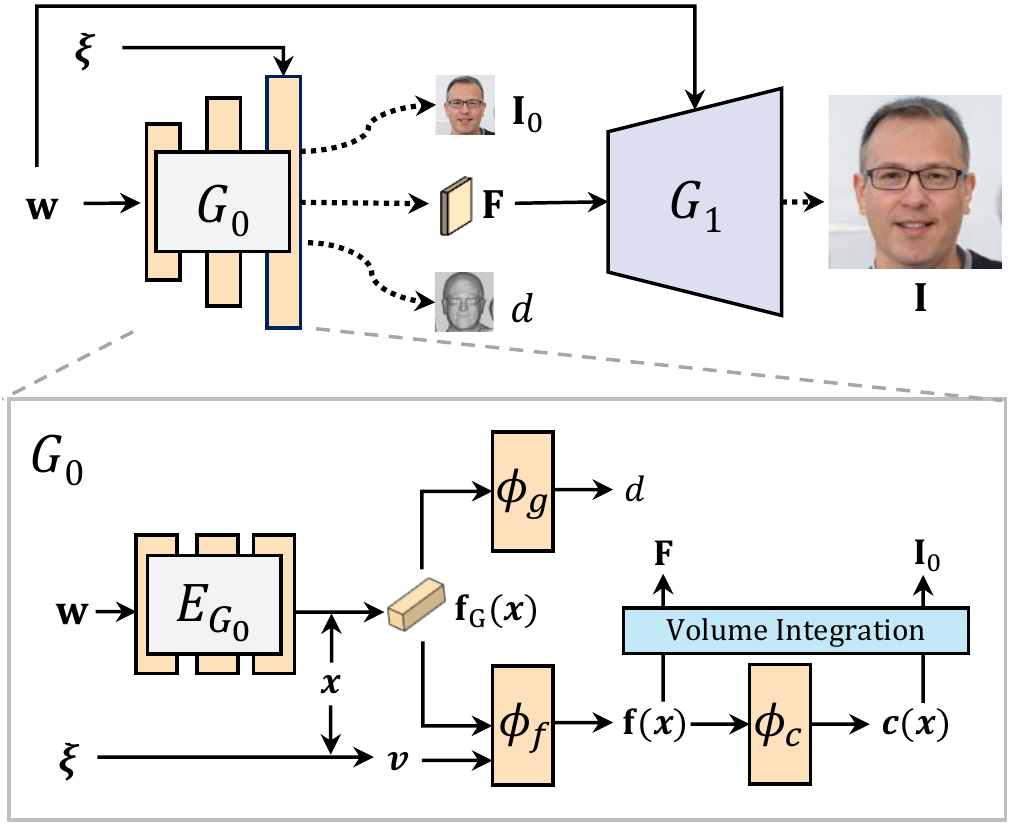}
\caption{\textbf{StyleSDF.} 
Given a sampled latent code $\wcode$ and a camera pose $\campose$, StyleSDF generates object SDF $\dist$ to depict the shape and the corresponding face image $\image$.}
\label{fig:datasetsample}
\end{figure}

\heading{Hybrid 3D-aware Generation}
To achieve high-resolution novel view synthesis, hybrid 3D-aware generator~\cite{GIRAFFE,gu2021stylenerf,Chan2021EG3D,orel2021stylesdf} is proposed. It is a cascade model
%
$\generator=\rendererG\circ\decoderG$ composed of a NeRF-based renderer $\rendererG$~\cite{Chan2021piGANPI} and a 2D super-resolution network $\decoderG$, as shown in Fig.~\ref{fig:datasetsample}.
Both $G_0$ and $G_1$ follow the style-based architecture~\cite{karras2019style,karras2020analyzing} to accept a latent code $\wcode$ to control the style of the generated object.
During generation, $G_0$ captures the underlying geometry with the full control of $\wcode$ and camera pose $\campose$, and renders a low resolution image $ \image_{\rendererflag}$ and an intermediate feature map $\mathbf{F}$.
Then, $\decoderG$ further upsamples $\mathbf{F}$ to obtain a high-resolution image $\image$ with added high-frequency details.

Among them, StyleSDF~\cite{orel2021stylesdf} introduces signed distance function (SDF) to serve as a proxy for the density function $\sigma(\point)$ used for the volume rendering in NeRF. 
Specifically, StyleSDF uses $\rendererG$ to predict the distance $\dist(\point)=\rendererG(\wcode,\point)$ between the query point $\point$ and the shape surface, where the density function $\sigma(\point)$ can be transformed from $\dist(\point)$ for NeRF~\cite{mildenhall2020nerf} to render.
The incorporation of SDF leads to higher-quality geometry in terms of expressiveness view-consistency and clear definition of the surface.
StyleSDF also enjoys the flexible style control for semantic editing as in StyleGAN~\cite{karras2019style}. 
Therefore, in this paper we mainly use StyleSDF as the base model for GAN inversion study.
Note that our method is not limited to StyleSDF and could be easily extended to other style-based 3D GAN variations.

\section{E3DGE}
An effective 3D GAN inversion shall be capable of 
\textbf{1)} reconstructing plausible 3D shape given single-view input,
\textbf{2)} maintaining high-fidelity texture,
and \textbf{3)} allowing view-consistent semantic edits.
To achieve these goals,
we propose the \nickname{} framework with three novel
components:
In Sec.~\ref{sec:global}, we leverage 3D GAN to generate
pseudo 2D-3D paired samples for 3D supervisions, and train an inversion encoder $\encoderGlobal$ to estimate the latent of plausible 3D shapes from a 2D image;
In Sec.~\ref{sec:local}, we train a local encoder $\encoderLocal$ to extract pixel-aligned features to enrich texture details for high-fidelity inversion; Finally, Sec.~\ref{sec:hybrid} introduces a hybrid alignment module for view-consistent semantic editing.

\subsection{Self-supervised Inversion Learning}
\label{sec:global}
%
In this section, we propose to mitigate the lack of large-scale high-quality 2D-3D paired datasets by retrofitting pre-trained 3D GANs to provide pseudo samples for training our inversion encoder.
We demonstrate the model trained from pseudo samples can rival and even outperform the methods learned from real data on the 3D GAN inversion task. We detail the process as follows.

\heading{Global Encoder for 3D GAN Inversion}
With the style-based $G$, we build our encoder $\encoderGlobal$ based on pSp~\cite{richardson2020encoding} for inversion. Given a target image $\image$, $\encoderGlobal$ predicts its latent code $\hat{\wcode}=\encoderGlobal(\image)$. 
Given the corresponding camera pose $\campose$, the reconstructed image is obtained by $\tilde{\image}=G(\hat{\wcode},\campose)$ to approximate $\image$. In addition, we would like its 3D shape predicted by $\rendererG$ to be plausible enough.

\heading{Distill 3D GANs as 3D Supervisions}
Different compositions of shape and texture could lead to identical 2D-rendered images. 
%
3D supervision is needed to alleviate such shape-texture ambiguity.
%
In the lack of large-scale high-quality 2D-3D paired samples, we 
formulate GAN Inversion as a \emph{self-training} task, 
where samples synthesized from itself are leveraged
to boost the reconstruction fidelity in both 2D and 3D domains.
%

As shown in Fig~\ref{fig:datasetsample}, we synthesize paired 3D shape information $\geo$ and 2D image $\image$ from latent code $\wcode$ and
camera pose $\campose$ using $G$ to train $\encoderGlobal$. To extract the 3D shape information $\geo$ of each synthetic shape, we first sample a point set $\gP=\{\surfaceset,\additionalset\}$ where $\surfaceset$ and $\additionalset$ contain points sampled from the surface and around the surface, respectively. Then, we calculate the geometry descriptor $\dist_i$ and $\normal_i$ for each 3D point $\point_i \in \gP$, and $\geo$ is defined as the set of geometry descriptors of all 3D point in $\gP$:
\begin{equation}
\begin{split}
    &
    \geo = \{ 
    \{{\dist}_{i}, {\normal}_{i}\}_{i=1}^{\setsize{\gP}} \mid \\
    &
    \point_i \in \mathcal{P},
    {\dist}_i = \rendererG(\wcode, \point_i), 
    {\normal}_{i}=\nabla_{\point_i}{\dist_{i}} 
    \},
\end{split}
\end{equation}
where ${\dist}_{i}$ is the distance from $\point_i$ to the shape surface and ${\normal}_{i}$ is the surface normal defined by the gradient of the distance w.r.t. $\point_i$.
Note our method is not limited to the SDF-based shape representation and can be easily extended to radiance-based methods~\cite{Chan2021piGANPI,pan2021shadegan,Chan2021EG3D}.
Moreover, given different camera poses, we can generate a diverse 2D-3D dataset to help alleviate the shape-texture ambiguity, \ie, for each shape $\geo$, various images $\image=\generator(\wcode,\campose)$ can be rendered by randomly sampling $\campose$ from a predefined pose distribution $p_{\campose}$.
Finally, we define $\datasetsample = \{\geo, \campose,\image\}$ as a training sample for $\encoderGlobal$.


%

\begin{figure}[t]
    \centering 
    \includegraphics[width=\linewidth]{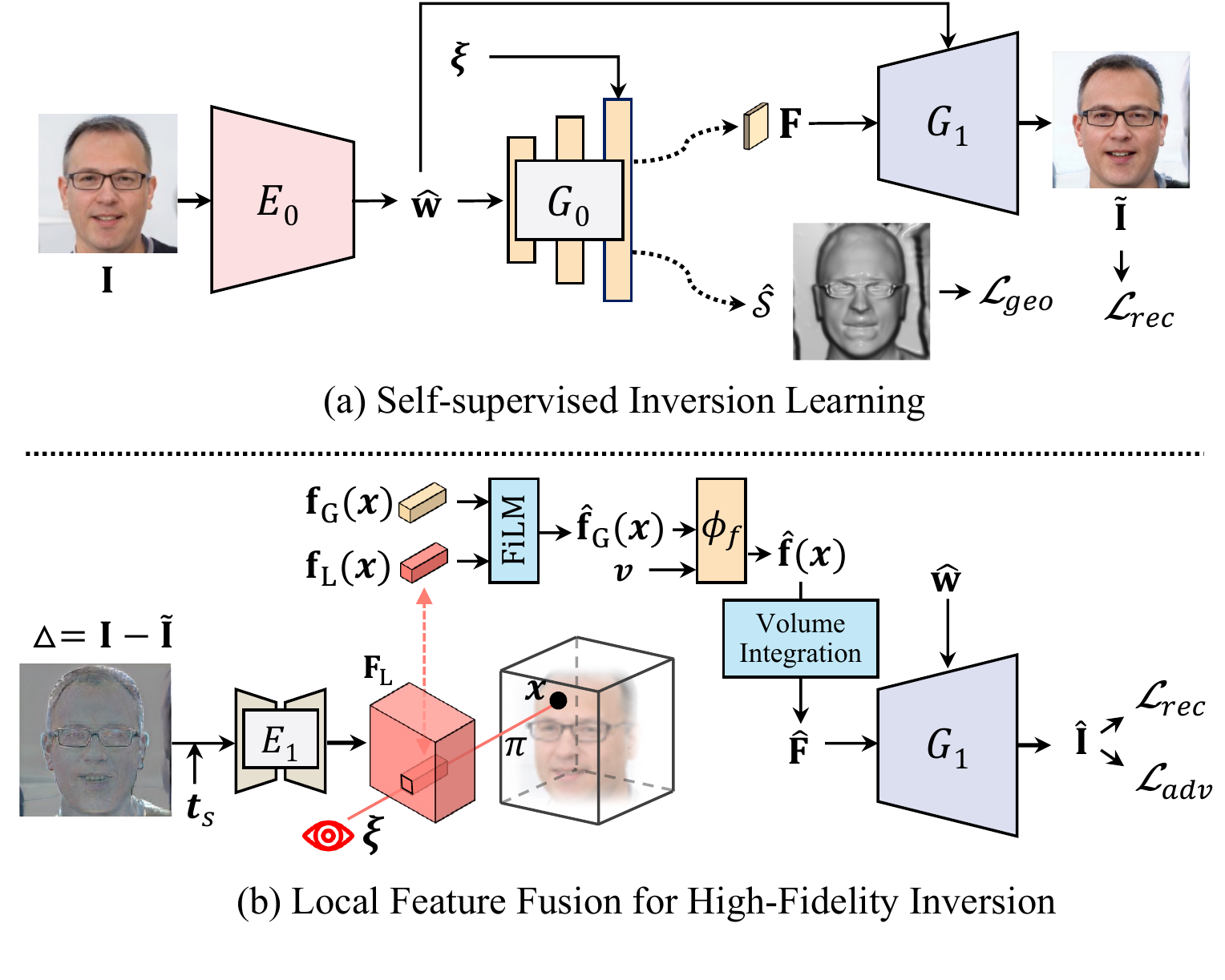}
\caption{\textbf{E3DGE for 3D GAN inversion}. (a) We augment the training of the encoder $\encoderGlobal$ with 3D supervision $\loss_{{geo}}$ for plausible 3D shape prediction. (b) We augment the representation capacity of the global latent code $\hat\wcode$ with local point-dependent latent feature $\mathbf{f}_\text{L}$ for high-fidelity texture reconstruction.}
\label{fig:stage1-2}
\end{figure}

\heading{3D GAN-Supervised Training}
As shown in Fig.~\ref{fig:stage1-2} (a),
given a training sample $\datasetsample$,
the forward process is represented as:
\begin{align}
{\hat{\wcode}} &= \encoderGlobal(\image) \\
\{ \tilde{\image}, \hat{\geo}\} &= \generator({\hat{\wcode}}, \campose, {\mathcal{P}})
\end{align}
where ${\hat{\wcode}}$ is the estimated latent code
and $\hat{\geo} = \{ \{ {\hat{\dist}}_{i}, {\hat{\normal}}_{i}\}_{i=1}^{\setsize{\gP}}  \mid \point_i \in \mathcal{P} \}$ is the estimated 3D shape information conditioned on $\tilde{\wcode}$ and $\mathcal{P}$.

To achieve 3D supervision, 
we would like the estimated $\hat{\geo}$ to approximate the ground truth $\geo$. Specifically, for points over the surface, their distances and normal are both considered while for points around the surface, we only supervise their distance following~\cite{park_deepsdf_2019,Alldieck2022PhotorealisticM3}, leading to geometry loss:
%
\begin{align}
\label{eq:geo}
    \loss_{geo}^{\set{O}} &= \mathbb{E}_{\datasetsample}\Bigg[
    \frac{1}{\setsize{\surfaceset}} \sum_{i=1}^{\setsize{\surfaceset}} \lossweight_{g_1}  
    |\hat{\dist}_{i}| + \lossweight_{g_2} \| \hat{\normal}_{i} - \normal_{i} \|_1 \Bigg] \\
    \loss_{geo}^{\set{F}} &= \mathbb{E}_{\datasetsample}\Bigg[
    \frac{1}{\setsize{\additionalset}} \sum_{i=1}^{\setsize{\additionalset}} \lossweight_{g_3} 
    |\hat{\dist}_{i} - \dist_{i}| \Bigg]  \\
    \loss_{geo} &= \loss_{geo}^{\set{O}} + \loss_{geo}^{\set{F}},
\end{align}
where $\lambda$s are loss weights and $\dist_{i}=0$ for points over the surface.
We also impose code reconstruction loss $\loss_{code}=\| \hat{\wcode} - \wcode \|_2$ to regularize the learning
and 2D supervisions $\loss_{rec}$ to minimize the reconstruction error between $\tilde{\image}$ and $\image$ as in pSp~\cite{richardson2020encoding}. 
The overall loss is $\loss = \loss_{geo} + \loss_{code} + \loss_{rec}$.
%
%

\begin{figure}[t]
    \centering 
    \includegraphics[width=1\linewidth]{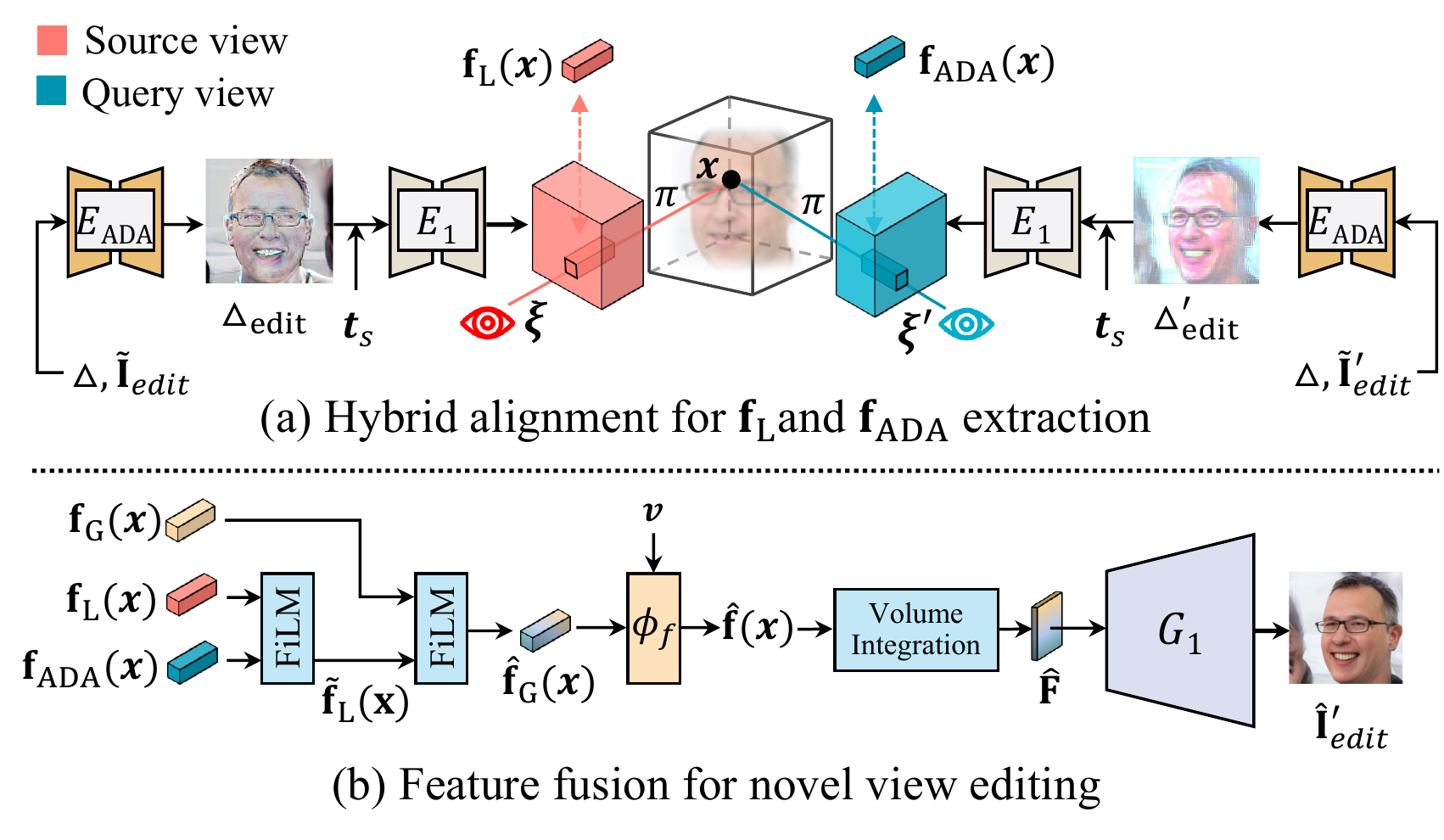}
\caption{\textbf{Hybrid alignment for high-quality editing.} 
Given code prediction $\hat{\wcode}$ from encoder $\encoderGlobal$ pre-trained in stage-$\RN{1}$,
we aim to generate high-quality view synthesis over the edited code $\hat{\wcode}_\text{edit}$.
In (a), the local details ${\residual}$ along with the target edited image $\image_\text{edit}^{\prime}$ and depth map $\bm{t}_s(\hat\wcode, \campose)$ are sent to pre-trained $\encoderADA$ to predict aligned residual ${\residual}^\prime_\text{edit}$.
The original aligned residual ${\residual}$ along with the 2D auxiliary residual ${\residual}^\prime_\text{edit}$ are processed by $\encoderLocal$ to recover latent maps $\mathbf{F}_\text{L}$ and $\mathbf{F}_\text{ADA}$ for later fusion. 
In (b), the extracted features $\mathbf{f}_\text{L}(\point)$ and $\mathbf{f}_\text{ADA}(\point)$ are first fused together with a FiLM layer,
and the fused result $\hat{\mathbf{f}}_\text{L}(\point)$ further serve as conditions to modulate the global feature $\mathbf{f}_\text{G}(\point)$.
The final modulated feature $\hat{\mathbf{f}}(\point)$ contains complete information, globally and locally.
The volume integrated $\hat{\mathbf{F}}$ is sent to $\decoderG$ for high-resolution synthesis.
}
\label{fig:stage3}
\end{figure}
\subsection{Local Features for High-Fidelity Inversion}\label{sec:local}

%
To facilitate introductions in the following sections, we first take a look at the details of StyleSDF. 
As shown in Fig.~\ref{fig:datasetsample}, $\rendererG$ can be further divided into four parts: 
a $8$-layer MLP encoder $E_{\rendererG}$, a SDF decoder $\phi_g$, a feature decoder $\phi_f$ and a color decoder $\phi_c$. 
$E_{\rendererG}$ extracts a global feature $\mathbf{f}_{\text{G}}(\point)=E_{\rendererG}(\point, \wcode)$.
Based on $\mathbf{f}_{\text{G}}$, $\phi_g$ and $\phi_f$ compute SDF  $\dist(\point)=\phi_g(\mathbf{f}_{\text{G}}(\point))$ and the last-layer feature $\mathbf{f}(\point, \view)= \phi_f(\mathbf{f}_{\text{G}}(\point), \view)$ of $\rendererG$, respectively. $\mathbf{f}$ could be directly transformed to color $\bm{c}(\point,\view)=\phi_c(\mathbf{f}(\point, \view))$ or being volume integrated to $\mathbf{F}$ and sent to $\decoderG$ for high resolution synthesis. For simplicity, we will omit $\view$ in the following.

%

\heading{Local Feature for Detailed Textures}
The global latent code $\hat{\wcode}$ is a compact representation of the predicted scene.
However, previous works~\cite{chan2022glean,wang2021HFGI} have validated that a low-dimensional latent code discards high-frequency spatial details and fails to reconstruct high-fidelity outputs.
This phenomenon becomes more severe when lifting the 2D image to a 3D scene, which contains exponentially more information. 
Inspired by recent progress in few-shot 3D reconstruction~\cite{saito2019pifu,saito2020pifuhd,yu2021pixelnerf,xiu2022icon,Alldieck2022PhotorealisticM3,Wang2021IBRNetLM,Chibane2021StereoRF},
we propose to make up for the lost information by introducing pixel-aligned (local) features.
As shown in Fig.~\ref{fig:stage1-2} (b), rather than conditioning all 3D points with the same latent code $\hat\wcode$,
we augment the representation capacity with local latent codes $\mathbf{f}_{\text{L}}$ that is dependent on each point $\point$.
We introduce a local hourglass~\cite{Newell2016StackedHN} encoder $\encoderLocal$ to predict a residual feature map $\mathbf{F}_\text{L}$ based on the reconstruction residue ${\residual}=\image-\tilde{\image}$,
\begin{equation}\label{eq:local_feat_3d}
    \mathbf{F}_\text{L} = \encoderLocal({\residual}, \bm{t}_s(\hat\wcode, \campose)),
\end{equation}
where $\bm{t}_s(\hat\wcode, \campose)$ is the depth map of the scene derived from the SDF to serve as 3D context information. Then, the local latent code of a point $\point$ is its corresponding value in $\mathbf{F}_\text{L}$:
\begin{equation}\label{eq:local_feat_3d2}
    \mathbf{f}_{\text{L}}(\point)=\textbf{F}_\text{L}(\pi(\point)) \oplus \textbf{PE}(\point),
\end{equation}
where $\pi$ maps the 3D point $\point$ to its corresponding pixel coordinate on 2D feature map $\mathbf{F}_\text{L}$. Since in 3D scenes, points along a ray will be projected to the same coordinate on the 2D plane, to differentiate these points, we additionally concatenate their positional encoding $\textbf{PE}(\point)$~\cite{mildenhall2020nerf} in Eq.~(\ref{eq:local_feat_3d2}).
In this way, the local feature $\mathbf{f}_{\text{L}}$ only encodes the residual information at the projected position $\pi(\point)$ but is also capable of determining where the residual information lies in the 3D scene, as well as inpainting the occluded areas along the ray.

Finally, we fuse the local latent code $\mathbf{f}_{\text{L}}(\point)$ with the global latent code $\mathbf{f}_{\text{G}}(\point)=E_{\rendererG}(\point, \hat\wcode)$ to 
supplement the missing high-frequency details.  
Specifically, the feature fusion is based on Feature-wise Linear Modulation (FiLM)~\cite{Perez2018FiLMVR}. As shown in Fig.~\ref{fig:stage1-2}, $\mathbf{f}_{\text{L}}(\point)$ is fed into two MLP layers to obtain the scale and bias modulation parameters $\mathbf{f}^{\bm\gamma}_{\text{L}}(\point)$ and $\mathbf{f}^{\bm\beta}_{\text{L}}(\point)$. Then we modulate $\mathbf{f}_{\text{G}}(\point)$ with FiLM 
\begin{equation}\label{eq:modulation}
  \hat{\mathbf{f}}_{\text{G}}(\point) = \text{FiLM}(\mathbf{f}_{\text{G}}(\point),\mathbf{f}_{\text{L}}(\point))={\mathbf{f}}_{\text{L}}^{\bm\gamma}(\point) \cdot {\mathbf{f}}_{\text{G}}(\point) +  {\mathbf{f}}_{\text{L}}^{\bm\beta}(\point).\nonumber
\end{equation}
The fused $\hat{\mathbf{f}}_{\text{G}}(\point)$ is volume integrated to $\hat{\mathbf{F}}$ and the final high-fidelity reconstructed image can be obtained as $\hat\image=\decoderG(\hat{\mathbf{F}})$.

Note that through point projection $\pi$, the reconstruction with local prior is not limited to the original view, and naturally works for novel views. However, for views with severe occlusions or additional editing, the residual features may not fully align with the scene, leading to a failed feature fusion.
We will address this issue in the next subsection with our hybrid feature alignment.

\subsection{Hybrid Alignment for High-Quality Editing}\label{sec:hybrid}
Though we achieve high-fidelity reconstruction with the aforementioned designs,
there is a trade-off between the \emph{input view} reconstruction quality and \emph{novel view} editing performance.
We first analyze the reasons behind and propose a hybrid alignment module to address this issue.

\heading{Reconstruction Editing Trade-off}
Given an input image $\image$ with paired reconstruction $\tilde{\image}$ and residual map ${\residual}$ extracted from the input view $\campose$ with the aforementioned method. 
First, at test time when the input image is edited $\tilde\image_\text{edit}$ or query view ${\campose^{\prime}} \neq \campose$, 
the residual map no longer aligns and is likely to result in wrong predictions.
Second, if we supervise the models to reconstruct the input itself,
the learned features are \emph{regressive} rather than \emph{generative} since all prediction areas are visible in the inputs.
With these two challenges, 
though the model could yield perfect reconstruction at training,
it would result in noticeable performance degradation when rendering from novel views at test time.

\heading{Hybrid Alignment for High-Quality Editing}
To address the first challenge,
we propose to infer aligned features with a 2D-3D hybrid alignment.
Specifically,
given edited latent code $\hat{\wcode}_\text{edit}$,
the initial novel-view edited image $\tilde\image^{\prime}_\text{edit}=\rendererG(\hat{\wcode}_\text{edit}, {\campose^{\prime}})$ is misaligned with ${\residual}$.
Inspired by HFGI~\cite{wang2021HFGI}, we leverage a 2D alignment module $\encoder_{\text{ADA}}$ to address the misalignment. 
As shown in Fig.~\ref{fig:stage3} (a),
we first obtain  ${\residual}_\text{edit}=\encoder_{\text{ADA}}({\residual},\rendererG(\hat{\wcode}_\text{edit}, {\campose}))$, transform it to residual feature map ${\mathbf{F}_\text{L}^\text{edit}}$ via Eq.~(\ref{eq:local_feat_3d}) and retrieve the view-consistent 3D local feature ${\mathbf{f}}_{\text{L}}$ via Eq.~(\ref{eq:local_feat_3d2}).
However, to render the high-quality edited image $\hat\image_{edit}^{\prime}$ from novel view $\campose^{\prime}$,
${\mathbf{F}_\text{L}^\text{edit}}$ might still suffer from occlusion due to large pose variations.
To the end, we propose a hybrid alignment to further refine ${\mathbf{F}_\text{L}^\text{edit}}$ with 2D aligned feature from $\encoder_{\text{ADA}}$. 
Specifically,
we align a 2D residue ${\residual}^{\prime}_\text{edit}=\encoder_{\text{ADA}}({\residual}, \tilde{\image}^{\prime}_{\text{edit}})$ 
and retrieve its corresponding ${\mathbf{f}}_{\text{ADA}}$ with $\encoderLocal$, which fills the occlusion in a 2D manner but lacks 3D consistency. 
To marry the best of both,
as shown in in Fig~\ref{fig:stage3} (b),
we modulate ${\mathbf{f}}_{\text{L}}$ with ${\mathbf{f}}_{\text{ADA}}$,
\begin{equation}\label{eq:refine_fusion_modulation}
  \tilde{\mathbf{f}}_\text{L}(\point) = \text{FiLM}(\mathbf{f}_{\text{L}}(\point),\mathbf{f}_{\text{ADA}}(\point))\text{,}
\end{equation}
and further fuse $\tilde{\mathbf{f}}_\text{L}$ with $\mathbf{f}_\text{G}(\point)$ for final prediction,
\begin{equation}\label{eq:global_modulation}
  \hat{\mathbf{f}}(\point) =
  {\text{FiLM}(\mathbf{f}_{\text{G}}(\point),\tilde{\mathbf{f}}_\text{L}(\point))\text{,}}
\end{equation}
\noindent 
where $\hat{\mathbf{f}}(\point)$ is then integrated to $\hat{\mathbf{F}}$ for rendering the final novel-view edited image $\hat{\image}_{edit}^{\prime}=\decoderG(\hat{\mathbf{F}})$.

\heading{Novel View Training for Coherent View Synthesis} 
To address the second challenge and enforce the model to learn generative features,
during training, we sample two views $\campose_1$ and $\campose_2$ for each style code $\wcode$, and render the corresponding images $\image^{\campose_1}$ and $\image^{\campose_2}$. 
Then, we train the models to reconstruct plausible novel views, \ie, $\generator(\encoder(\image^{\campose_1}), \campose_2) \approx \image^{\campose_2}$ and $\generator(\encoder(\image^{\campose_2}),\campose_1) \approx \image^{\campose_1}$. 
This training strategy facilitates a high-quality view synthesis over edited scenes.


\vspace{-1mm}
\section{Experiments} 
\label{sec:experiment}


\begin{table*}[h!]
\centering
\small
\renewcommand{\arraystretch}{1.1}
\caption{\textbf{Quantitative comparison for inversion quality on faces.}}
\label{tab:2DMetrics}
\begin{tabular}{l@{\hspace{3mm}}c@{\hspace{3mm}}c@{\hspace{3mm}}c@{\hspace{3mm}}c@{\hspace{3mm}}|c@{\hspace{3mm}}c@{\hspace{3mm}}c@{\hspace{3mm}}c@{}}

\toprule
 & \multicolumn{4}{c|}{Source View Reconstruction} & \multicolumn{4}{c}{Novel View Reconstruction} \\
\cmidrule[0.5pt]{2-9}
Method & MAE  $\downarrow$ &  SSIM $\uparrow$ &   LPIPS $\downarrow$          & Similarity $\uparrow$       & MAE  $\downarrow$ &  SSIM $\uparrow$ &   LPIPS $\downarrow$            & Similarity $\uparrow$       \\
                                 \midrule
pSp$_\text{StyleSDF}$   & .150 $\pm$ .032 & .696 $\pm$ .048 & .270 $\pm$ .059 & .498 $\pm$ .099 & .235 $\pm$ .010 & .604 $\pm$ .011 & .358 $\pm$ .048 & .513 $\pm$ .041               \\
e4e$_\text{StyleSDF}$  & .174 $\pm$ .049 & .669 $\pm$ .049 & .226 $\pm$ .063 & .252 $\pm$ .107 & .237 $\pm$ .014 & .597 $\pm$ .011 & .341 $\pm$ .063 & .271 $\pm$ .060 \\
\NICKNAME{}     & \textbf{.097 $\pm$ .008} & \textbf{.780 $\pm$ .016} &\textbf{ .128 $\pm$ .017} &\textbf{ .883 $\pm$ .017} &.\textbf{173 $\pm$ .008} & \textbf{.710 $\pm$ .010 }& \textbf{.154 $\pm$ .016} & \textbf{.903 $\pm$ .021} \\
\bottomrule
\end{tabular}
\end{table*}

\heading{Datasets} We mainly focus on the human face domain and use both 2D and 3D datasets for extensive evaluation.
To examine 2D reconstruction quality, we adopt CelebA-HQ~\cite{karras2018progressive,CelebAMask-HQ} dataset for source view reconstruction.
To further evaluate novel view reconstruction performance, we synthesize $500$ trajectory videos from a pretrained generator as a proxy test set.
For attribute editing, we adopt InterfaceGAN~\cite{Shen2020InterFaceGANIT} and Talk2Edit~\cite{jiang2021talkedit} to search for the editing directions.
To evaluate 3D shape reconstruction quality, we use NoW benchmark~\cite{RingNet:CVPR:2019} that provides a rich variety of face images with ground-truth 3D scans.
The 3D GANs are pre-trained on FFHQ~\cite{karras2019style}.
Note that our method does not rely on any external 3D data during the training process.

\heading{Implementation Details} 
For all the encoder models,
we adopt Adam optimizer with a learning rate of $5e-5$  to
train the models on 4 NVIDIA Tesla V100 GPUs, with a resolution of $256^2$, 
batch size of 24, and 16 samples along a ray for the recommended $200K$ iterations.
Following~\cite{saito2019pifu}, we filter our invisible 3D points when training from a certain view.
Code, dataset, and all pre-trained models will be made publicly available.
More details are included in the supplementary material. 

\subsection{Evaluation}
\subsubsection{Quantitative Evaluation} 
Since existing baselines are trained on StyleGAN~\cite{karras_analyzing_2020} and could be directly applied, 
for comparison, we implement two canonical encoder-based GAN inversion approaches on StyleSDF~\cite{orel2021stylesdf}, \ie, pSp~\cite{richardson2020encoding} and e4e~\cite{tov2021designing}, which stress reconstruction and editing quality respectively.

\heading{2D Reconstruction}
For 2D evaluation, we report inversion performance for both source view reconstruction and novel view reconstruction in Tab~\ref{tab:2DMetrics}.
For source view reconstruction, the metrics are calculated on the $2,825$ images from CelebA-HQ test set~\cite{CelebAMask-HQ}.
For novel view reconstruction, the metrics are averaged from $500$ videos generated from pre-trained 3D GANs,
each with $250$ frames covering ellipsoid camera poses trajectory. 
For each video, we randomly pick one image as source view input and the remaining images as ground truths with labeled poses as query views.
In this way, we could extensively evaluate the view synthesis ability under occlusions and varied input viewpoints.
%
Our approach substantially outperforms encoder-based baselines in terms of reconstruction quality in both source view and target view.
We include the comparison in the supplementary material and show that our method is considerably faster than optimization-based methods during inference.

\begin{table}[h]
\centering
\caption{Performance of 3D face reconstruction on NoW~\cite{RingNet:CVPR:2019}.}
\label{tab:3Dmetrics_now}
\begin{tabular}{l@{\hspace{2mm}}c@{\hspace{2mm}}c@{\hspace{2mm}}c@{\hspace{2mm}}c@{\hspace{2mm}}}
\toprule
Methods                                          & {Prior Type} & Median$\downarrow$       & Mean$\downarrow$         & Std                            \\ 
\midrule
3DMM-CNN~\cite{tran2017regressing}                                         & 3DMM                           & 1.84                           & 2.33                           & 2.05                           \\
PRNet~\cite{feng2018prn}                                            & 3DMM                           & 1.50                           & 1.98                           & 1.88                           \\
RingNet~\cite{RingNet:CVPR:2019}                                          & FLAME                          & 1.21                           & 1.54                           & 1.31                           \\
3DDFA-V2                                         & 3DMM                           & 1.23                           & 1.57                           & 1.39                           \\
DECA~\cite{DECA:Siggraph2021}                                             & FLAME                          & 1.09                           & 1.38                           & 1.18                           \\ 
\midrule
Wu et al.~\cite{wu2020unsupervised}  & Model Free                     & 2.64                           & 3.29                           & 2.86                           \\ 
\midrule
Ours                                             & 3D GAN                         & 1.70 & 2.08 & 1.67            \\ 
\bottomrule
\end{tabular}
\vspace{-0.25cm}
\end{table}

\heading{3D Reconstruction}
\begin{figure}[h!]
    \centering 
    \includegraphics[width=1\linewidth]{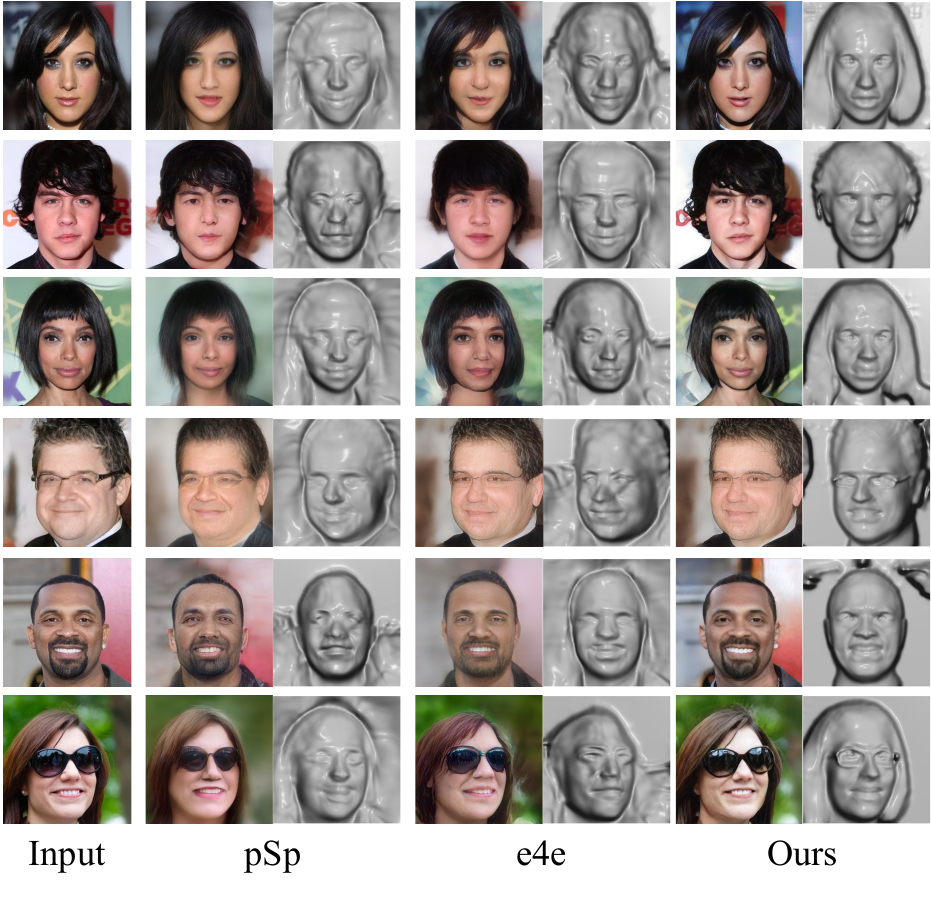}
    \vspace{-0.5cm}
\caption{\textbf{Qualitative comparisons on face inversions.}}
\label{fig:reconstruction-face}
\end{figure}
We report the 3D face reconstruction performance on NoW benchmark test set in Tab.~\ref{tab:3Dmetrics_now}.
Our method surpasses purely model-free method~\cite{wu2020unsupervised} and shows competitive performance compared with methods designed for 3D face reconstruction using basic models, \eg, 3DMM~\cite{blanz1999morphable} and FLAME~\cite{FLAME:SiggraphAsia2017}.
Note that as discussed in Wu~\etal~\cite{wu2020unsupervised}, NoW benchmark is designed for model-based reconstruction methods and inherently put model-free approaches at a disadvantage.
Therefore, our method could serve as a reference for fair quantitative evaluation comparisons of future model-free methods.

%

\begin{figure*}[h!]
    \centering 
    \includegraphics[width=1\linewidth]{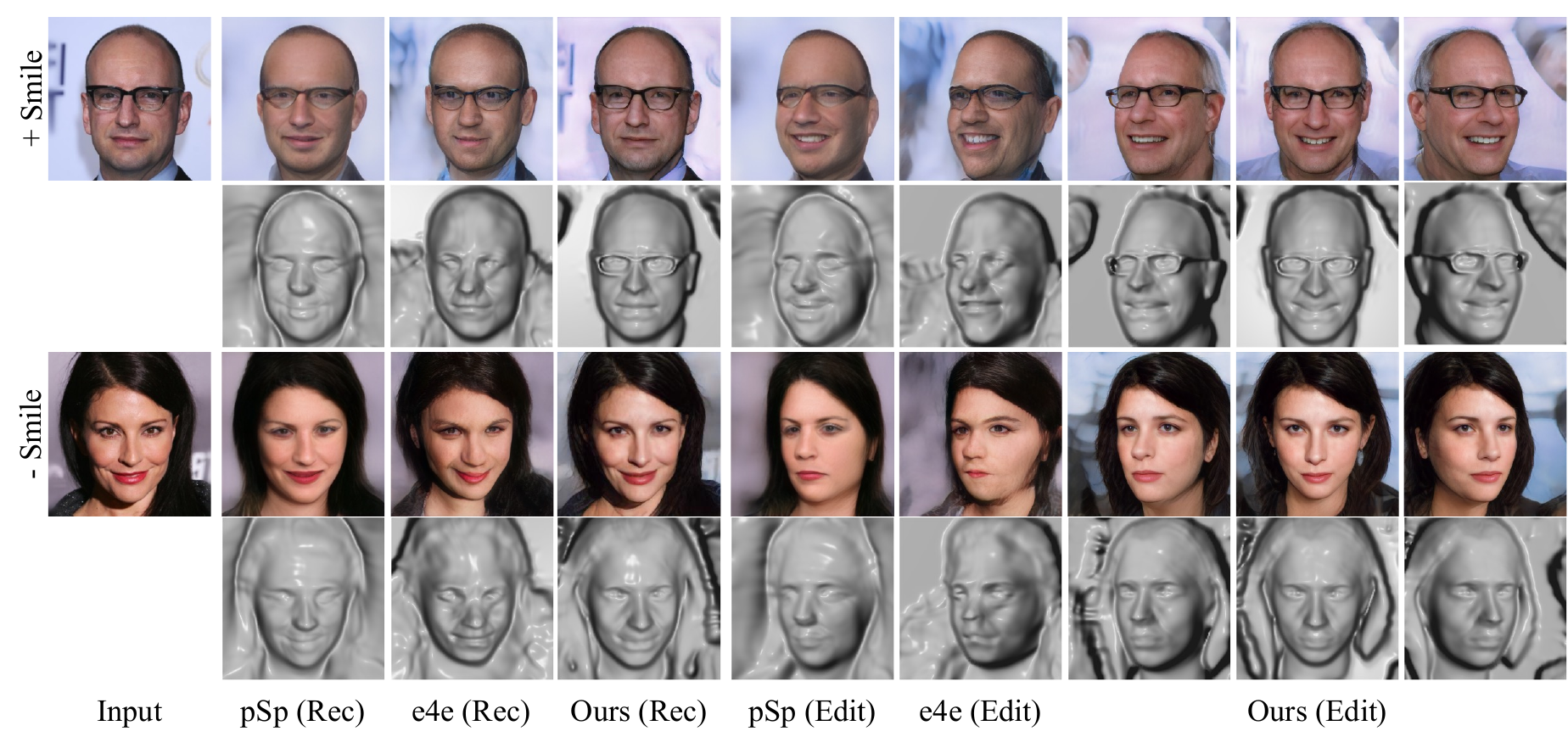}
\caption{\textbf{Qualitative comparisons on face inversion and editing under novel views.}}
\label{fig:editing-face}
\vskip -0.2cm
\end{figure*}

\subsubsection{Qualitative Evaluation}
\heading{Reconstruction} 
We show reconstruction performance in Fig.~\ref{fig:reconstruction-face}. 
\emph{Geometry-wise}, the baseline models without explicit 3D supervisions tend to generate implausible intermediate shapes, \eg, e4e predictions of rows $(3,6)$ and pSp predictions of rows $(2,5)$. Besides, their reconstruction is not close to the ``ground truth", and the reconstructed surface lacks details.
Our method successfully regularizes the intermediate 3D shapes and generates plausible results with surface details and a more complete structure. For instance, in rows $4$ and $6$, our method reconstructs 3D eyeglasses in which the baselines fail.
Corresponding metrics in Tab.~\ref{tab:abla:3d} also validate the usefulness of the direct geometry supervisions and loss designs.
\emph{Texture-wise},
existing methods generate distorted results and suffer artifacts and identity change.
In contrast, with pixel-aligned features incorporated,
our method is more robust with high-fidelity results. 
In particular, our method captures more details and preserves the identity of different input viewpoints.
For example, in row $1-3$, our method accurately reconstructs the hair, and in row $5$, the beard.

\heading{Editing}
We include the editing results in Fig.~\ref{fig:editing-face} and choose the ``Smile" attribute for editing. Beyond plausible shape reconstruction with high-fidelity texture inversion, 
in-view synthesis over edited results, our method consistently generates high-quality edited renderings in terms of view consistency, details conservation, and identity preservation.
Compared with our method,
the baselines either fail to render intact identity (column $5$) or generate visually plausible shapes (column $6$).

\begin{figure}[h]
    \centering 
    \includegraphics[width=1\linewidth]{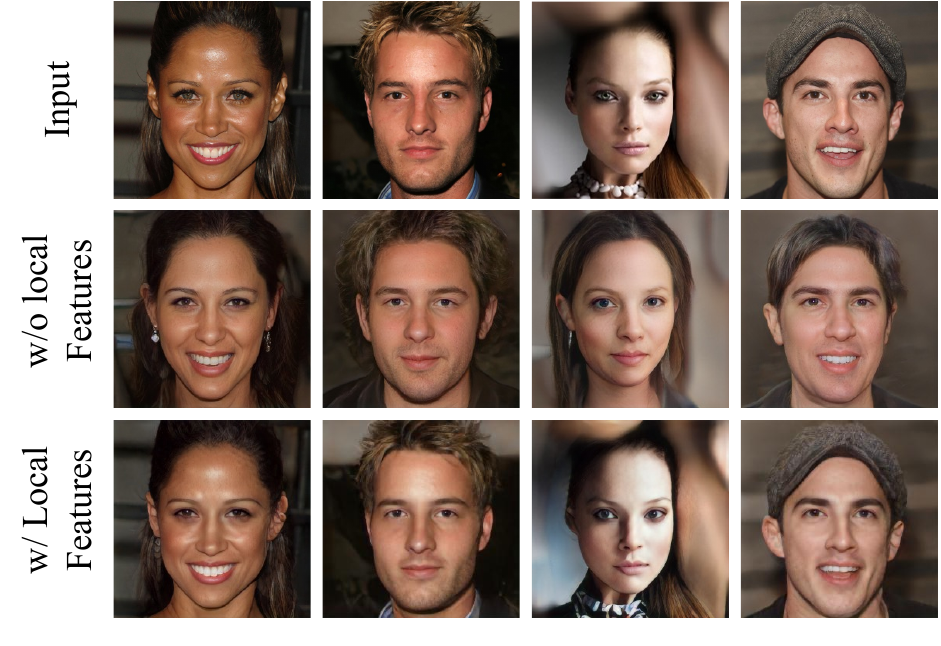}
\caption{\textbf{Ablation of Local Features.} Our method with pixel-aligned features shows photorealistic reconstructions..}
\label{fig:abla:globallocal}
\end{figure}

\begin{table*}[h!]
\centering
\small
\renewcommand{\arraystretch}{1.1}
\caption{\textbf{Ablations of Local Features and Hybrid Fusion.} 
Our local-global model design with hybrid alignment achieves the balance of high-quality reconstruction and view synthesis.}
\label{tab:abla:local-hybrid}
\begin{tabular}{l@{\hspace{3mm}}c@{\hspace{3mm}}c@{\hspace{3mm}}c@{\hspace{3mm}}c@{\hspace{3mm}}|c@{\hspace{3mm}}c@{\hspace{3mm}}c@{\hspace{3mm}}c@{}}

\toprule
& \multicolumn{4}{c|}{Source View Reconstruction} & \multicolumn{4}{c}{Novel View Reconstruction} \\
\cmidrule[0.5pt]{2-9}
Ablation Settings & MAE  $\downarrow$ &  SSIM $\uparrow$ &   LPIPS $\downarrow$          & Similarity $\uparrow$       & MAE  $\downarrow$ &  SSIM $\uparrow$ &   LPIPS $\downarrow$            & Similarity $\uparrow$       \\
                                 \midrule
Synthetic Training & .245 $\pm$ .024 & .634 $\pm$ .019 & .333 $\pm$ .029 & .369 $\pm$ .056 & .241 $\pm$ .011 & .594 $\pm$ .008 & .366 $\pm$ .059 & .770 $\pm$ .026 \\
$+$Local Features     & \textbf{.074 $\pm$ .007} & \textbf{.811 $\pm$ .015 }& \textbf{.075 $\pm$ .010} & \textbf{.953 $\pm$ .006 }&  .282 $\pm$ .103              &    .571 $\pm$ 0.056            &        .511 $\pm$ 0.031        &  .608 $\pm$ .123              \\
\midrule
$+$3D Alignment         & .102 $\pm$ .009 & .772 $\pm$ .015 & .119 $\pm$ .016 & .818 $\pm$ .029 & {.133 $\pm$ .011} & {.709 $\pm$ .022} & \textbf{.130 $\pm$ .021} & .901 $\pm$ .011 \\
$+$2D Alignment         & .098 $\pm$ .005 & .774 $\pm$ .038 & .140 $\pm$ .040 & .900 $\pm$ .032 & .178 $\pm$ .007 & .656 $\pm$ .009 & .178 $\pm$ .012 & \textbf{.904 $\pm$ .018}               \\
Hybrid Alignment     & .097 $\pm$ .008 & .780 $\pm$ .016 & .128 $\pm$ .017 & .883 $\pm$ .017 & \textbf{.131 $\pm$ .008} & \textbf{.710 $\pm$ .010} & .154 $\pm$ .016 & .903 $\pm$ .021 \\
\bottomrule
\end{tabular}
\vspace{-3mm}
\end{table*}
\subsection{Ablation Study}
\heading{Effect of 3D GAN as Supervisions}
We quantitatively validate the effects of 3D supervision in NoW Challenge validation set and report the corresponding metrics in Tab.~\ref{tab:abla:3d}.
For the results of fully synthetic dataset training (row $1$),
compared with the baseline method with a similar network (pSp), fully synthetic data training shows worse reconstruction metrics. We attribute this phenomenon to the domain gap between synthesized images and real images. However, our method shows surprisingly better performance over identity preservation in novel views ($0.77$ compared with $0.513$ of pSp$_\text{StyleSDF}$ and $0.271$ of e4e$_\text{StyleSDF}$, which we attribute to the well-aligned pose of synthetic corpus leads to less distortion in view synthesis.
\begin{table}[t]
\centering
\caption{Effect of 3D Supervisions.}
\vspace{-0.3cm}
\label{tab:abla:3d}
\begin{tabular}{lccc}
\toprule
Settings                                      & Median$\downarrow$ & Mean$\downarrow$ & Std \\ 
\midrule
pSp$_\text{StyleSDF}$                                   & 1.97 & 2.43 & 2.05 \\
e4e$_\text{StyleSDF}$                                    & 2.83 & 3.40 & 2.67 \\
\midrule
$+ \loss_{geo}^{\set{O}}$       & 1.75 & 2.11 & 1.72 \\
$+ \loss_{geo}^{\set{F}}$  & 1.71 & 2.09 & 1.70  \\
$+ \loss_{code}$      & 1.66 & 2.06 & 1.69 \\
\bottomrule
\end{tabular}
\vspace{-0.3cm}
\end{table}

\heading{Effect of Local Features}
As discussed before, the local features preserve the missing image details to facilitate high-fidelity reconstruction. To validate the effectiveness of local features in texture reconstructions, we show the inversion results in Fig.~\ref{fig:abla:globallocal}. 
With the proposed local-global fusion pipeline, our model captures more details and guarantees photorealistic reconstruction.
Quantitative results in Tab.~\ref{tab:abla:local-hybrid} also validate the effectiveness of local features in high-quality inversion.
The results on the video trajectories also show that without delicate design, \eg novel-view training,
local features would fully collapse over novel view synthesis.

\begin{figure}[h]
    \centering 
    \includegraphics[width=1\linewidth]{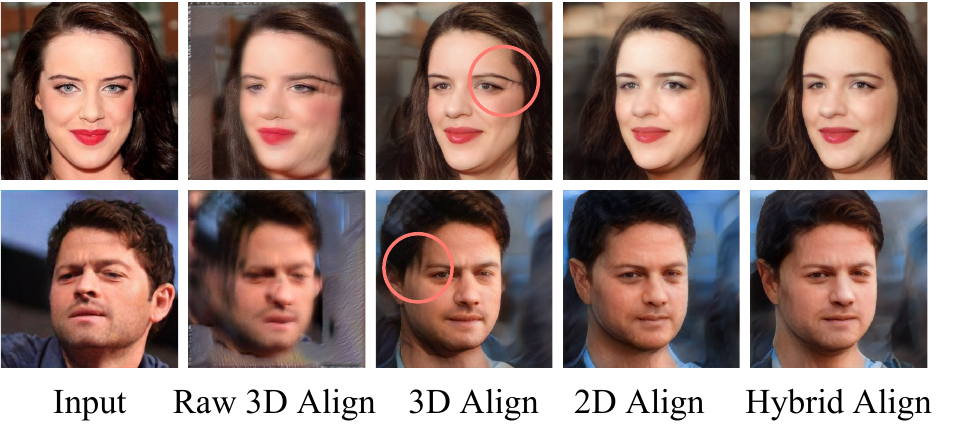}
    \vspace{-0.5cm}
\caption{\textbf{Ablation of Hybrid Alignment.} From left to right, we show the novel view synthesis of raw 3D-aligned features w/wo novel-view training, synthesis achieved using 2D-aligned features, and the final hybrid features. 3D-aligned features are view-consistent but suffer from occlusions (circled), while 2D features are visually plausible but lack some details (\eg, hair color). Our hybrid fused results share the best of both.}
\label{fig:abla:fusion}
\end{figure}

\heading{Effect of Hybrid Alignment}
We show the view synthesis achieved by different alignment methods in Fig.~\ref{fig:abla:fusion}.
To quantitatively analyze the effect of hybrid alignment, in Tab.~\ref{tab:abla:local-hybrid} we evaluate the model performance of 3D alignment and 2D alignment individually. For both ablations, novel-view training is enabled. As shown here, the 3D alignment model shows better view consistency in video prediction measured by reconstruction metrics, and the 2D alignment model shows better identity preservation. 
The hybrid alignment model marries the best of both and also enables semantic editing and yields better reconstruction performance on the video predictions.

%


\section{Conclusion and Discussions}
We propose a novel 3D GAN
inversion framework E3DGE for 3D face reconstruction and editing.
We marry the benefits of both self-supervised global prior and pixel-aligned local prior for high-quality shape and texture reconstruction.
A hybrid alignment that bridges the best of 2D and 3D features is further proposed for view-consistent editing. 
Benefiting from the overall system design, the proposed method has advantages in terms of both high fidelity and editability. 
As a pioneer attempt in this direction,
we believe this work opens a new line of research direction and will inspire future works on 3D GAN inversion, few-shot 3D reconstruction and 3D-aware learning from 2D images. 

\vspace{0.1cm}
\heading{Limitations and Future Work}
The proposed method suffers data bias introduced by the synthetic data. As the synthetic data lacks complex details and pose variations compared with real-world data, our method trained with it tends to generate simple background and fail on extreme poses. 
Special attentions should be paid to data bias to avoid social impact to under represented minorities.
A future direction is to leverage real data for semi-supervised training. 
Another future direction is to leverage the hyper-network for efficient local feature incorporation to alleviate the extra computational cost of the 2D alignment module.
Finally, we would explore the potentials of our framework on other 3D GANs and shapes beyond human face and other editing methods uniquely designed for 3D GANs.




\clearpage


\clearpage

{\small
\bibliographystyle{ieee_fullname}
\bibliography{bibs/cvpr23.bib}
}


\end{document}


\title{Self-Supervised Geometry-Aware Encoder for Style-Based 3D GAN Inversion\\\vspace{1em}
Supplementary Material\vspace{-1em}}

\maketitle








\newpage
\appendix

\section{Background}
Since recent 3D-aware image generative models are all based on neural implicit representations, especially NeRF~\cite{mildenhall2020nerf}, here we briefly introduce the NeRF-based 3D representation and more StyleSDF details for clarification.

\heading{NeRF-based 3D Representation}
NeRF~\cite{mildenhall2020nerf} proposed an implicit 3D representation for novel view synthesis. 
Specifically, NeRF defines a scene as $\{\bm{c},\sigma\} = F_{\Phi}(\point, \view)$, 
where $\point$ is the query point, $\view$ is the viewing direction from camera origin to $\point$, $\bm{c}$ is the emitted radiance (RGB value), 
$\sigma$ is the volume density. 
To query the RGB value $C(\bm{r})$ of a point on a ray $\bm{r}(t)=\bm{o}+t\view$ shoot from the 3D coordinate origin $\bm{o}$, 
we have the volume rendering formulation,
\begin{equation}
\label{eq:volum_render}
C(\bm{r}) = \int_{t_n}^{t_f}T(t)\sigma(\bm{r}(t))\bm{c}(\bm{r}(t), \bm{v})dt,
\end{equation}
where $T(t)=\text{exp}(-\int_{t_n}^{t}\sigma(\bm{r}(s))ds)$
is the accumulated transmittance along the ray $\bm{r}$ from $t_n$ to $t$. $t_n$ and $t_f$ denote the near and far bounds. 

\heading{More StyleSDF Details}
In hybrid 3D generation~\cite{orel2021stylesdf,Chan2021EG3D,gu2021stylenerf}, the intermediate feature map is calculated by replacing the color $\bm{c}$ with feature $\mathbf{f}$ from $\phi_{f}$, namely $\mathbf{F}(\mathbf{r}) = \int_{t_n}^{t_f}T(t)\sigma(\mathbf{r}(t))\mathbf{f}(\mathbf{r}(t),\bm{v})dt$.
In StyleSDF, the Sigmoid activation function $\sigma$ is replaced by $\sigma(\point) = K_{\alpha} \left( \dist(\point) \right) =  \text{Sigmoid}\left(-\dist(\point)/\alpha\right)/\alpha$,
where $\alpha$ is a learned parameter that controls the tightness of the density around the surface boundary.

\heading{Notation Table}
For clarity, we include the notations used in the proposed method in Tab.~\ref{tab:supp:notation}.

\section{Implementation Details}
\subsection{More Methods Details}

\heading{Surface Point Sampling in Self-supervised Inversion Learning}
In Sec.~{\color{red}4.1} of the main paper, to extract the 3D shape information $\geo$ of each synthetic shape, we first sample a point set $\gP=\{\surfaceset,\additionalset\}$ where $\surfaceset$ and $\additionalset$ contain points sampled from the surface and around the surface, respectively.
To get points over the surface $\surfaceset$ for training, 
for efficiency, we directly reuse the intermediate results to render $\image_{\rendererflag}$ to calculate the surface. 
Specially, to sample point set $\set{O}$ we replace the color $\mathbf{c}$ as the coordinates $\point$ of points along a ray in Eq.~(\ref{eq:volum_render})  and approximate the 3D coordinates of surface, namely 
$\bm{t}_s(\wcode, \campose) = \int_{t_n}^{t_f}T(t, \wcode)\sigma(\bm{r}(t),\wcode)t~dt$.
In this way, we get $B \times H \times W$ surface points for training in each iteration, where $B$ stands for batch size and $H \times W$ stands for the resolution to render 3D consistent images, \eg, $64 \times 64$.
To sample point set $\set{F}$, we add Gaussian offset to each of the calculated surface points $\set{O}$. Specifically, we adopt Gaussian distribution $\gN(0, (r/4)^{2})$ where $r$ is the radius of the scene. In this way, points falling within $4$ standard deviations would cover $95.44\%$ of the whole 3D space. Following PIFu~\cite{pifuSHNMKL19}, we also uniformly sample $0.5 \times B \times H \times W$ points within the whole 3D space defined. The overall quantity of the point set surface is $\setsize{\set{F}} = 1.5 \times B \times H \times W$.
We find this sampling strategy avoids overfitting and yields better performance.

\heading{Training Details of High-Fidelity Inversion With Local Features}
In Sec.~{\color{red}4.2} of the main paper, we train a local encoder $\encoderLocal$ to extract pixel-aligned features to enrich texture details for high-fidelity inversion.
The network architecture of $\encoderLocal$ is identical to that of PIFu~\cite{pifuSHNMKL19}, which is a stacked hourglass network with residual connections. The input residual map resolution is $256\times256$, and the output $64\times64$ resolution feature map.
${\mathbf{f}}_{\text{L}} \in \gR^{256}$ is bilinearly interpolated from feature map $ \mathbf{F}_\text{L}$ at the projected position $\pi(\point)$.
As shown in Fig.~\ref{fig:supp:FiLM},
we implement the FiLM layer~\cite{perez2018film} with two MLP residual blocks~\cite{yu2021pixelnerf}, which outputs $\alpha$ and $\beta$ for modulation, respectively.
We use the identical learning rate and optimizer to train $\encoderLocal$.

\heading{Novel-View Training Details}
For novel-view training for coherent view synthesis in Sec.~{\color{red}4.3} of the main paper, in each training iteration with batch size $n$, rather than sampling $n$ different latent codes $\{\zcode_{i}\}_{i=1}^{n}$, we halve the number of identical latent codes $\{\zcode_{i}\}_{i=1}^{n/2}$ while double the rendered images for each latent code $\{ \image_{i}^{\campose_1},\image_{i}^{\campose_2} \}_{i=1}^{n/2}$
where $n$ is even.
Thus, we train the models to reconstruct plausible \emph{novel views}, \ie, $\generator(\encoder(\image_{i}^{\campose_1}), \campose_2) \approx \image_{i}^{\campose_2}$ and $\generator(\encoder(\image_{i}^{\campose_2}),\campose_1) \approx \image_{i}^{\campose_1}$. Since the paired-sampled images could serve as both inputs and ground truths, the effective batch size and training cost maintains the same.
To train 2D alignment model $\encoderADA$, we further regularize the predicted residual map $\hat{\residual}^{\campose_1} \approx \image^{\campose_1}-\image_{\rendererflag}^{\campose_1}$ with $\loss_{1}$ loss, where $\image_{\rendererflag}^{\campose_1}$ is corresponding renderer output low-resolution image and $\lambda_{1}=0.1$.
Note that we finetune pre-trained $\encoderADA$ from HFGI~\cite{wang2021HFGI} with novel-view training and no edited images are involved in the training time.

\heading{Curriculum Pose Sampling}
At the beginning of the training of the hybrid alignment in Sec.~{\color{red}4.3} of the main paper, large view changes will make the prediction of residual features and the inpainting of occlusion regions extremely difficult. As a result, our model is prone to blurry results. 
We attribute the reason to the ill-posed nature of rendering novel views given partial observations since the inpainted image is not unique.
To facilitate novel-view training,
we design a curriculum learning strategy~\cite{duan2020cdeepsdf} based on \emph{pose sampling difficulty}.
Implementation wise, 
given the camera pose distribution $\campose \sim {p_{\campose}}$ with mean $\mu$ and standard variance $\sigma$,
we fix the $\mu$ and scale the $\sigma$ with a weight $\alpha$ which is initially set to $0$ and gradually increases to $1$ as the training goes.
Intuitively,
when $\alpha=0$ the source view $\campose$ is identical to the query view $\campose^{\prime}$,
the training degrades to a regression task where the model shall reconstruct all the texture details to minimize the loss.
As the variance $\alpha \cdot \sigma$ increases,
the training becomes a conditional generation task to inpaint plausible and photo-realistic areas.
\subsection{More Experiments Details}
\heading{Training Details}
In this work,
we directly use the officially released pre-trained GAN models from StyleSDF.
In self-supervised shape inversion learning (Sec.~{\color{red}4.1}), due to GPU memory restriction, we sample $4$ shapes per GPU each iteration for training.
After $\encoderGlobal$ converged, we fix the network weights and only train the $\encoderLocal$ for high-fidelity inversion.
We train each stage for $50,000$ iterations, 
which costs 2 days on $4$ Tesla V100 GPUs.

\begin{figure}[t]
    \centering 
    \includegraphics[width=1\linewidth]{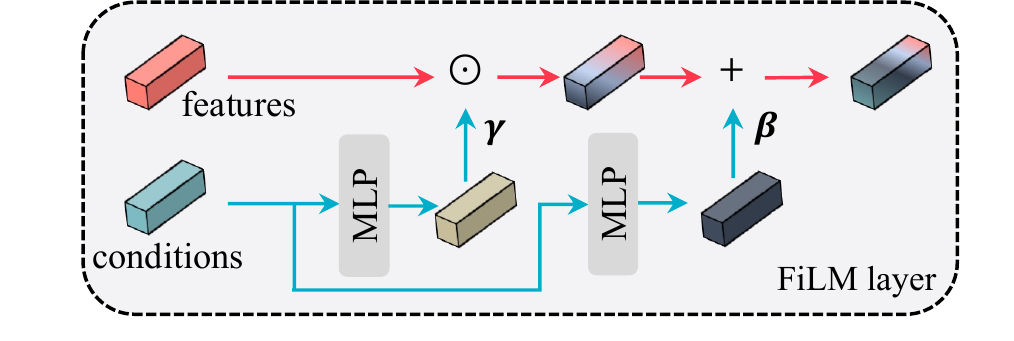}
\vspace{-5mm}
\caption{{FiLM Layer Architecture.}}
\label{fig:supp:FiLM}
\end{figure}

\heading{Network Architecture Details}
For $\encoderGlobal$,
a modified version of the pSp encoder~\cite{richardson2020encoding} is deployed here for a fair comparison with existing work.
Since $G_0$ and $G_1$ of StyleSDF have $9$ and $10$ latent codes, respectively, we introduce $9+10$ extra prediction heads to the pSp for the latent code prediction.
We observe that early layers of $\rendererG$ control the geometry of generated samples, and later $\rendererG$ layers as well as decoder generator $\decoderG$ control the texture and high-frequency details.
Thus, we adopt the early pSp feature map of resolution $32\times 32$ to predict latent code of $G_0$ for geometry control, and pSp feature map of resolution $64 \times 64$ to predict latent code of $G_0$ for texture control. 
We use the highest resolution feature map of pSp with resolution $128 \times 128$ to predict the latent code for $G_1$.
We show our FiLM layer implementation in Fig.~\ref{fig:supp:FiLM}, where the input features are modulated by the input conditions with predicted $\gamma, \text{and}~\beta$.
The MLP is implemented with the MLP residual block~\cite{yu2021pixelnerf}.

\heading{Editing}
For attribute editing, following previous works, we adopt vector-arithmetic~\cite{Radford2016UnsupervisedRL} based editing.
Specifically, a searched latent code vector paired with a certain attribute is weighted and added to the predicted code $\hat{\wcode}$.
To search for the meaningful editing directions on the 3D GAN used, we first sample $10,000$ images with paired latent codes from StyleSDF,
and then apply the face attribute predictor from Talk-to-Edit~\cite{jiang2021talkedit} to predict the corresponding attributes score. 
Based on the prediction, we apply SVM classifier from InterfaceGAN~\cite{Shen2020InterFaceGANIT} to search for the decision boundary. As in previous works~\cite{tov2021designing,richardson2020encoding}, we search for the editing latent code in the $\wspace$ space.

\heading{3D Face Reconstruction Evaluation Details}
We evaluate the reconstructed 3D meshes and compare them with the performance of several model-based reconstruction methods on NoW benchmark~\cite{RingNet:CVPR:2019}.
NoW benchmark~\cite{RingNet:CVPR:2019}, provides a test set of $1,702$ images of $80$ subjects and a ground-truth 3D scan per subject. 
These images are captured with a higher variety in facial expression, occlusion, and lighting and shall validate the generality of single-view reconstruction methods under real-world conditions.

To extract meshes for evaluation, we detect faces and crop the images using RetinaFace~\cite{serengil2021lightface} implemented by~\cite{facexlib} and obtain 3D mesh reconstructions from the depth maps predicted by our method trained on FFHQ pre-trained generator. We then use the evaluation protocol provided by the benchmark, which aligns the predicted meshes with the ground-truth meshes with a rigid transformation based on seven pre-defined keypoints and computes the scan-to-mesh distances. We obtain keypoints on our predicted meshes by applying a facial keypoint detector~\cite{Wang_2019_ICCV_awing} on the reconstructed canonical images. Following Unsup3D~\cite{wu2020unsupervised}, the average keypoints are used when the keypoint detector fails.

\heading{Video Trajectory Evaluation Details}
We sample $500$ trajectory videos with pre-trained FFHQ StyleSDF generator with an ellipsoid trajectory of size $250$ from official StyleSDF code, making a dataset of size $12,5000$. The evaluation code and dataset will be released.

\subsection{Losses}

\heading{Reconstruction Loss}
We briefly introduce the supervisions we adopt in image reconstructions in both training stages.
First, we utilize the pixel-wise $\loss_2$ loss,
\vspace{-1.15mm}
\begin{equation}
    \loss_{\text{2}}\left ( \image \right ) = || \image - \imagePredicted ||_2.
\end{equation}

\vspace{-1.15mm}
In addition, to learn perceptual similarities, we use the LPIPS~\cite{zhang2018perceptual} 
loss, which has been shown to better preserve image quality compared to the more standard perceptual loss:
\begin{equation}
    \loss_{\text{LPIPS}}\left ( \image \right ) = || F(\image) - F(\imagePredicted)||_2,
\end{equation}
where $F(\cdot)$ denotes the perceptual feature extractor.

Finally, a common challenge when handling the specific task of encoding facial images is the preservation of the input identity. To tackle this, we incorporate a dedicated recognition loss measuring the cosine similarity between the output image and its source, 
\vspace{-1.25mm}
\begin{equation}
    \loss_{\text{Similarity}}\left (\image \right ) = 1-\left \langle R(\image),R(E_g(\image)) \right \rangle ,
\end{equation}
where $R$ is the pretrained ArcFace~\cite{deng2019arcface} network.

In summary, the total loss function is defined as
\begin{equation*}
    \loss_{rec}(\image) = 
    \lossweight_1 \loss_2(\image) + 
    \lossweight_2 \loss_\mathrm{LPIPS}(\image) + 
    \lossweight_3 \loss_\mathrm{Similarity}(\image) \text{,}
\end{equation*}

\noindent
where we set $\lossweight_1=1$, $\lossweight_2=0.8$, $\lossweight_3=0.1$ as the defined loss weights. 
In $\encoderGlobal$ training,
we supervise images $\hat\image_{\rendererflag}, \hat\image_{\decoderflag}$ of both resolutions.
In $\encoderLocal$ training,
we only supervise the reconstruction of high-resolution images since the network weights to render $\hat\image_{\rendererflag}$ is fixed.
Here,
we also impose the non-saturating adversarial loss with R1 regularization~\cite{Mescheder2018WhichTM} to improve the naturalness of reconstructed images, which is defined as:
\begin{align}
\label{eq:ganloss}
    \loss_{{adv}} &= -\mathop{\mathbb{E}}[log(D(\imagePredicted))], \\
    \loss_D &= \mathop{\mathbb{E}}[log(D(\imagePredicted))] +  \mathop{\mathbb{E}}[log(1-D(\image))], \\
    \loss_{{R1}} &=  \lambda\|\nabla D({\imagePredicted}; \theta_D)\|_2, 
\end{align}
where $D$ is initialized with the pre-trained discriminator paired with the generator and $\theta_D$ is the corresponding parameters to optimize.
In summary, the overall loss is the weighted summation of of the loss functions described above:
\begin{equation}
    \loss = 
    \loss_{geo} + 
    \loss_{rec} +
    \lossweight_{adv} \loss_{{adv}} + 
    \lossweight_{D} \loss_{{D}} +
    \lossweight_{R1} \loss_{{R1}} \text{,}
\end{equation}

\noindent
where we set $\lossweight_{D}=\lossweight_{adv}=0.01$ and $\lossweight_{R1}=10$ in the experiments.

\begin{table*}[h!]
\centering
\caption{Notations used in the proposed method.}
\label{tab:supp:notation}
\begin{tabular}{ll}
\toprule
Notation                                      & Meaning \\ 
\midrule
$\hat{*}$                          & Final predictions                                                                 \\
$\tilde{*}$                        & Intermediate results                                                              \\
${*}^{\prime}$                     & Abbreviation of target view camera pose                                           \\
$\generator$                       & Generator                                                                         \\
$\rendererG$                       & Renderer Generator                                                                \\
$\decoderG$                        & SR Generator                                                                      \\
$\decoder$                         & Discriminator                                                                     \\
$\encoder$                         & Encoder                                                                           \\
$\encoderGlobal$                   & Encoder to predict global latent code                                             \\
$\encoderLocal$                    & Hourglass encoder to predict pixel-aligned local features.                        \\
$\encoderADA$                      & ADA (Adaptive Distortion Alignment) module                                        \\
$\wspace$                          & W space for style-based GAN                                                       \\
$\wcode$                           & Latent code sampled from W space                                                  \\
$\image$                           & Input image                                                                       \\
$\image_{\rendererflag}$           & Rendered image from renderer generator                                            \\
$\imageEdit$                       & Edited image                                                                      \\
$\hat\wcode$                       & Predicted latent code from $\encoderGlobal$                                       \\
$\lambda$                          & Loss weights                                                                      \\
{$\point$}                      & {3D point}                                                                      \\
$\gP$                              & Point set                                                                         \\
$\surfaceset$                      & Point set sampled from object surface                                             \\
$\additionalset$                   & Point set sampled near the surface or uniformly in the defined 3D space.              \\
$\dist$                            & Signed distance function                                                          \\
$\normal$                          & Normal for a point                                                                \\
$\mlpGeo$                          & MLP to predict geometry                                                           \\
$\mlpFeature$                      & MLP to predict view-dependent feature                                             \\
$\mlpTex$                          & MLP to predict color                                                              \\
$\view$                            & View direction                                                                    \\
$\datasetsample$                   & A synthetic data sample for training                                              \\
$\campose$                         & Source view camera pose                                                           \\
$\campose^{\prime}$                & Target view camera pose                                                           \\
$\residual$                        & Residual of predicted image and input image                                       \\
$\residual_\text{edit}$            & Residual paired with an edited image                                                 \\
$\residual^{\prime}_\text{edit}$   & Residual paired with an edited image rendered from target camera pose.               \\
$\pi(\point)$                      & Projection of 3D point $\point$ to source view                                            \\
$\oplus$                           & Concatenation                                                                     \\
$\textbf{PE}$                      & Positional Encoding                                                               \\
$\beta$, $\gamma$                  & Modulation signals for FiLM                                                       \\
$\mathbf{t}_{s}(\wcode, \campose)$ & Depth map for code $\wcode$ rendered from pose $\campose$                         \\
$\feature$                         & Feature map                                                                       \\
$\featureLocal$                    & Local feature map output from $\encoderLocal$                                     \\
$\hat{\feature}$                   & Modulated feature map for final prediction                                        \\
$\feature_\text{ADA}$              & Local feature map output from $\encoderLocal$ with $\encoderADA$ aligned residual \\
$\mathbf{f}_\text{G}$              & Global feature output from the generator.                                         \\
$\mathbf{f}_\text{L}$              & Local feature interpolated from $\featureLocal$                                   \\
$\mathbf{f}_\text{ADA}$            & Aligned feature interpolated from $\featureLocal$                                 \\
$\hat{\mathbf{f}}_\text{L}$        & Predicted local feature for final prediction                                      \\
\bottomrule
\end{tabular}
\end{table*}

\section{More Results}
\heading{Comparisons with Optimization-based Methods}
We include the comparisons with two canonical optimization-based methods here, namely SG2~\cite{karras2019style,abdal2019image2stylegan} which is initially proposed in StyleGAN~\cite{karras2019style} paper to project input image to the $\wspace$ space of the paired generator, and PTI~\cite{roich2021pivotal} which further finetune the generator weights to achieve high-fidelity inversion. 
We implement SG2 and PTI following the official implementations and tune the corresponding parameters for StyleSDF generator. 
For SG2, we optimize $450$ steps with learning rate $5e-3$, and for the pivotal tuning stage, we optimize $100$ steps with learning rate $5e-5$.
We will release all inversion-related code upon acceptance.

\begin{figure*}[h!]
    \centering 
    \includegraphics[width=1\linewidth]{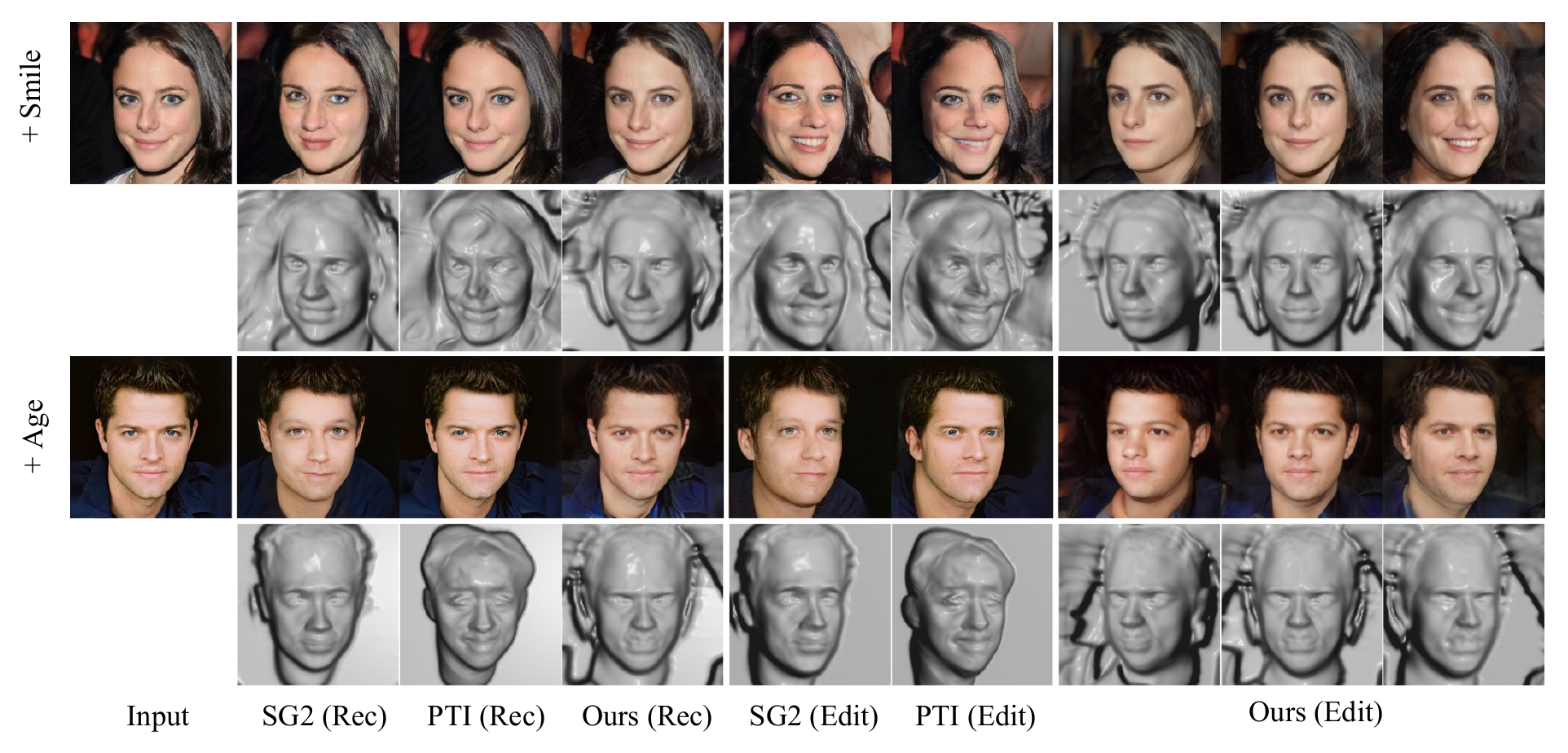}
\vspace{-6mm}
\caption{Visual comparisons on optimization-based methods. 'Rec' and 'Edit' represent reconstruction and editing, respectively.}
\label{fig:supp:editing:optimization}
\end{figure*}

We show the qualitative comparison in Fig.~\ref{fig:supp:editing:optimization}. As can be seen, SG2 could not reconstruct high-fidelity texture details but maintains a plausible intermediate shape inversion, due to the strong regularization of $\wspace$ space.
Though PTI could achieve photorealistic reconstruction, it still could not alleviate the shape-texture ambiguity, leaving the inverted shape distorted. 

We also include the quantitative comparisons in Tab.~\ref{tab:supp:optimization-based-comparison}. 
Specifically, for 2D inversion metrics, we inverse each image in the test set ($2,780$ CelebA-HQ images) with SG2 and PTI and calculate the reconstruction metrics as well as the inference time. 
For 3D inversion metrics, we adopt the NoW challenge validation set and reconstruct the corresponding depth mesh for $352$ identities.
As can be seen, SG2 cannot achieve high-fidelity reconstruction, and PTI could yield high-quality reconstruction at the cost of inference time and shape quality. Our proposed method achieves a balance of both and holds the merit of speedy inference, with only $0.19$ seconds needed to render an image from a novel view.

\begin{table*}[tp]
\centering
\small
\renewcommand{\arraystretch}{1.1}
\caption{Quantitative comparisons with optimization-based methods on faces.}
\label{tab:supp:optimization-based-comparison}
\begin{tabular}{l@{\hspace{3mm}}c@{\hspace{3mm}}c@{\hspace{3mm}}c@{\hspace{3mm}}c@{\hspace{3mm}}|c@{\hspace{3mm}}c@{\hspace{3mm}}c@{\hspace{3mm}}|c@{}}

\toprule
 & \multicolumn{4}{c|}{CelebA-HQ~\cite{CelebAMask-HQ}} & \multicolumn{3}{c|}{NoW Challenge~\cite{RingNet:CVPR:2019} Validation Set}  & Inference Time \\
\cmidrule[0.5pt]{2-9}
Method & MAE  $\downarrow$ &  SSIM $\uparrow$ &   LPIPS $\downarrow$          & Similarity $\uparrow$       & Median$\downarrow$       & Mean$\downarrow$         & Std  & Second (s)  $\downarrow$\\
                                 \midrule
SG2~\cite{karras2019style}   & .202 $\pm$ .063 & .650 $\pm$ .054 & .167 $\pm$ .046 & .219 $\pm$ .106 & 1.89 & 2.23 & 1.82   & 235s   \\
PTI~\cite{roich2021pivotal}  & \textbf{.062 $\pm$ .012} &  \textbf{.796 $\pm$ .017} & \textbf{.027 $\pm$ .005} & \textbf{.892 $\pm$ .009} &  2.86 & 3.54 & 3.01  & 265s \\
\midrule
\NICKNAME{}     & {.097 $\pm$ .008} & {.780 $\pm$ .016} &{ .128 $\pm$ .017} &{ .883 $\pm$ .017} & \textbf{1.66} & \textbf{2.06} & \textbf{1.69}  & \textbf{0.19s (Texture) / 0.81s (Shape)} \\
\bottomrule
\end{tabular}
\end{table*}

\begin{figure*}[h!]
    \centering 
    \includegraphics[width=1\linewidth]{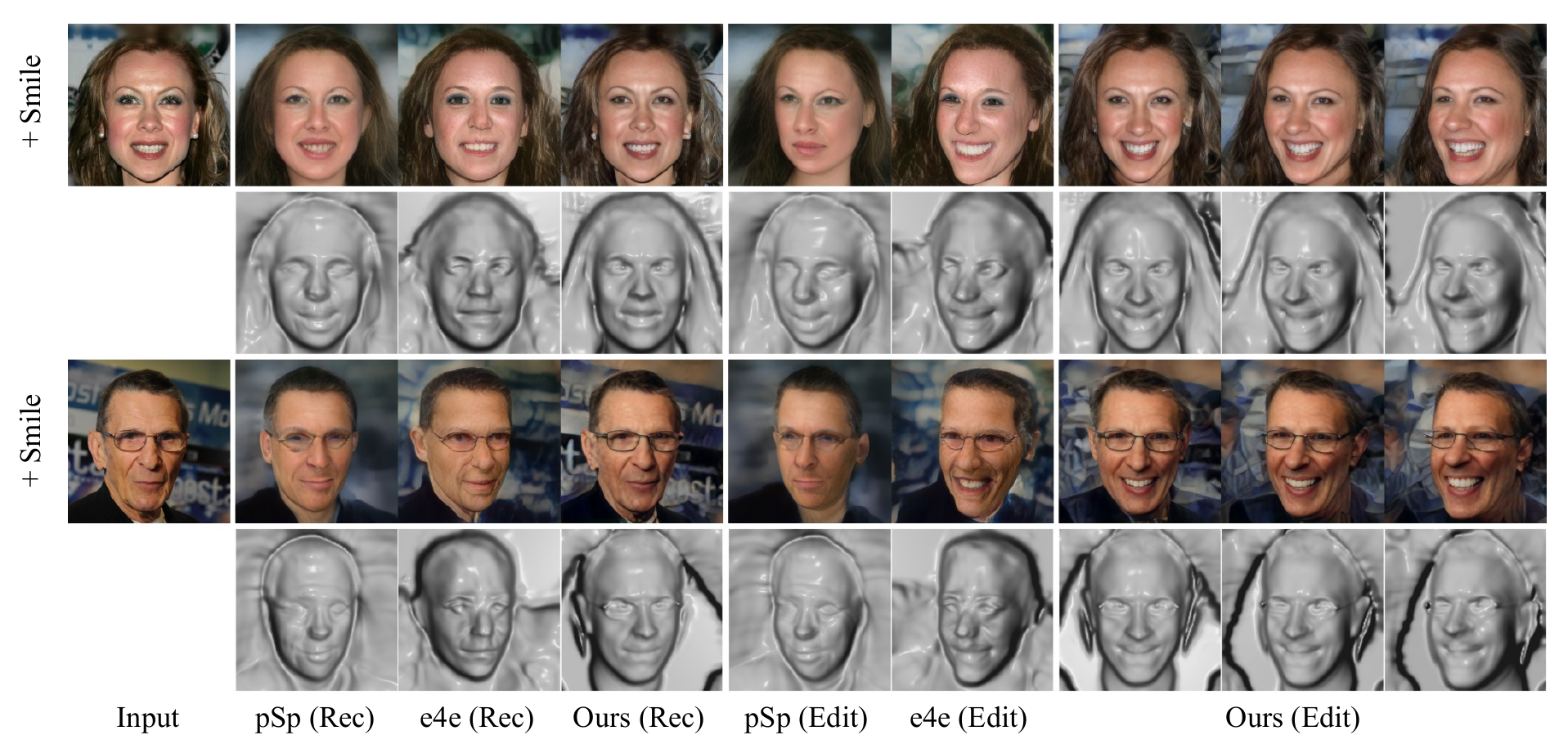}
\vspace{-6mm}
\caption{Visual comparisons on encoder-based methods. 'Rec' and 'Edit' represent reconstruction and editing, respectively.}
\label{fig:supp:editing:encoder}
\end{figure*}

\heading{More Comparisons with Encoder-based Methods}
Here, we include more comparisons with encoder-based methods in Fig.~\ref{fig:supp:editing:encoder}. Our method achieves consistently better performance compared to the baselines in terms of reconstruction fidelity and editing visual quality.

\begin{figure*}[tp]
    \centering 
    \includegraphics[width=1\linewidth]{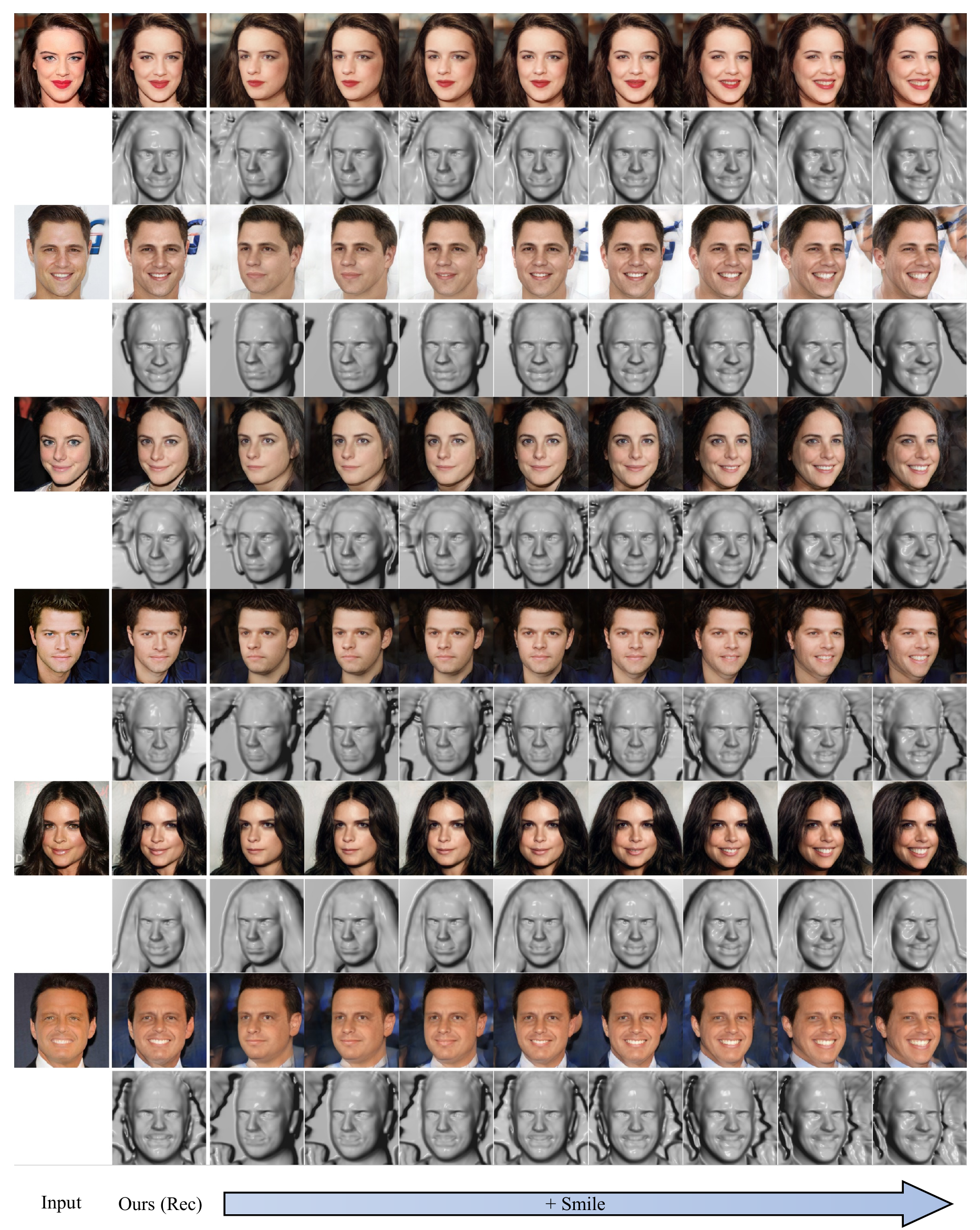}
\caption{Visual comparisons on face editing (Smile).}
\label{fig:supp:editing:smile}
\end{figure*}

\begin{figure*}[tp]
    \centering 
    \includegraphics[width=1\linewidth]{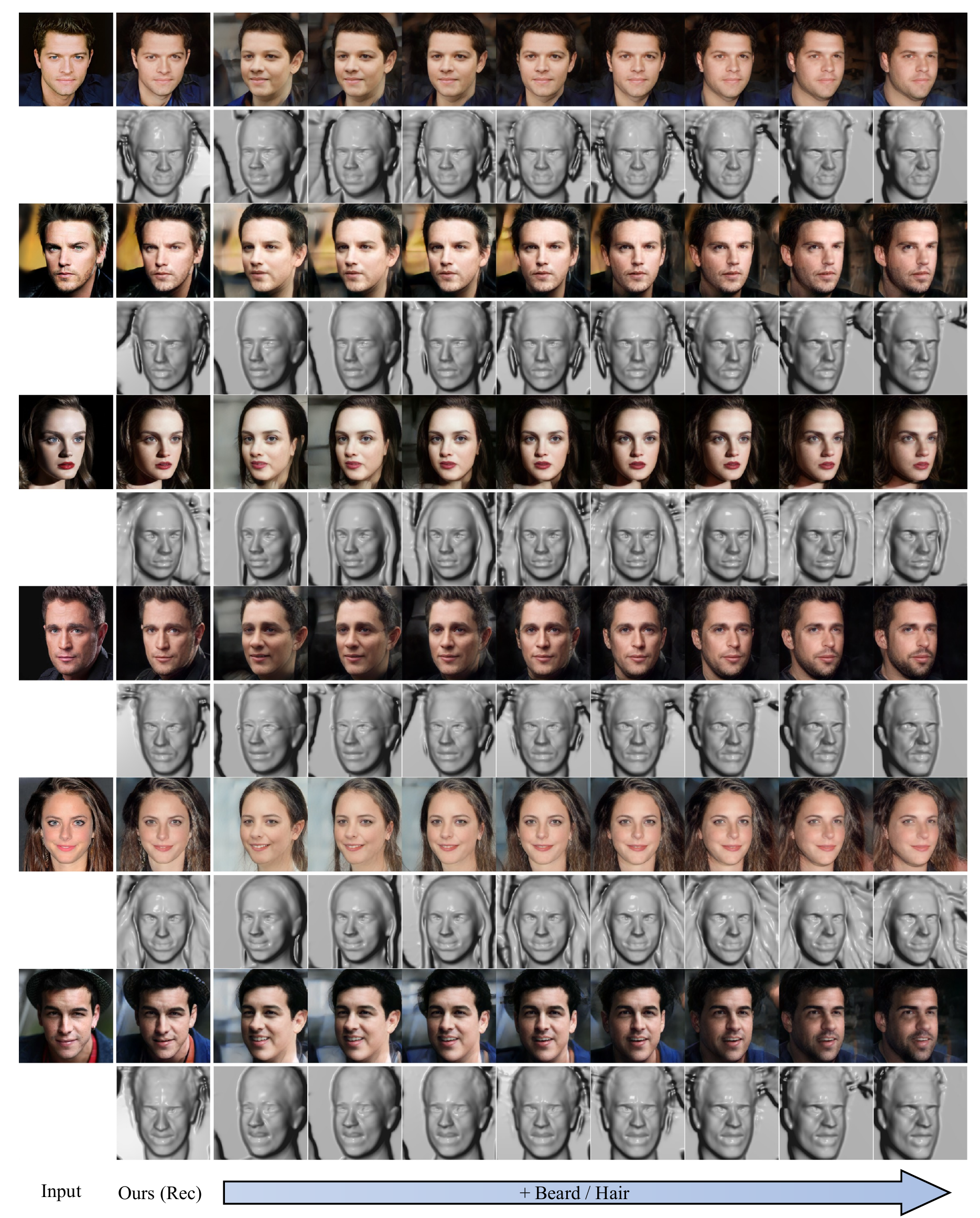}
\caption{Visual comparisons on face editing (Beard / Hair).}
\label{fig:supp:editing:beard}
\end{figure*}

\begin{figure*}[tp]
    \centering 
    \includegraphics[width=1\linewidth]{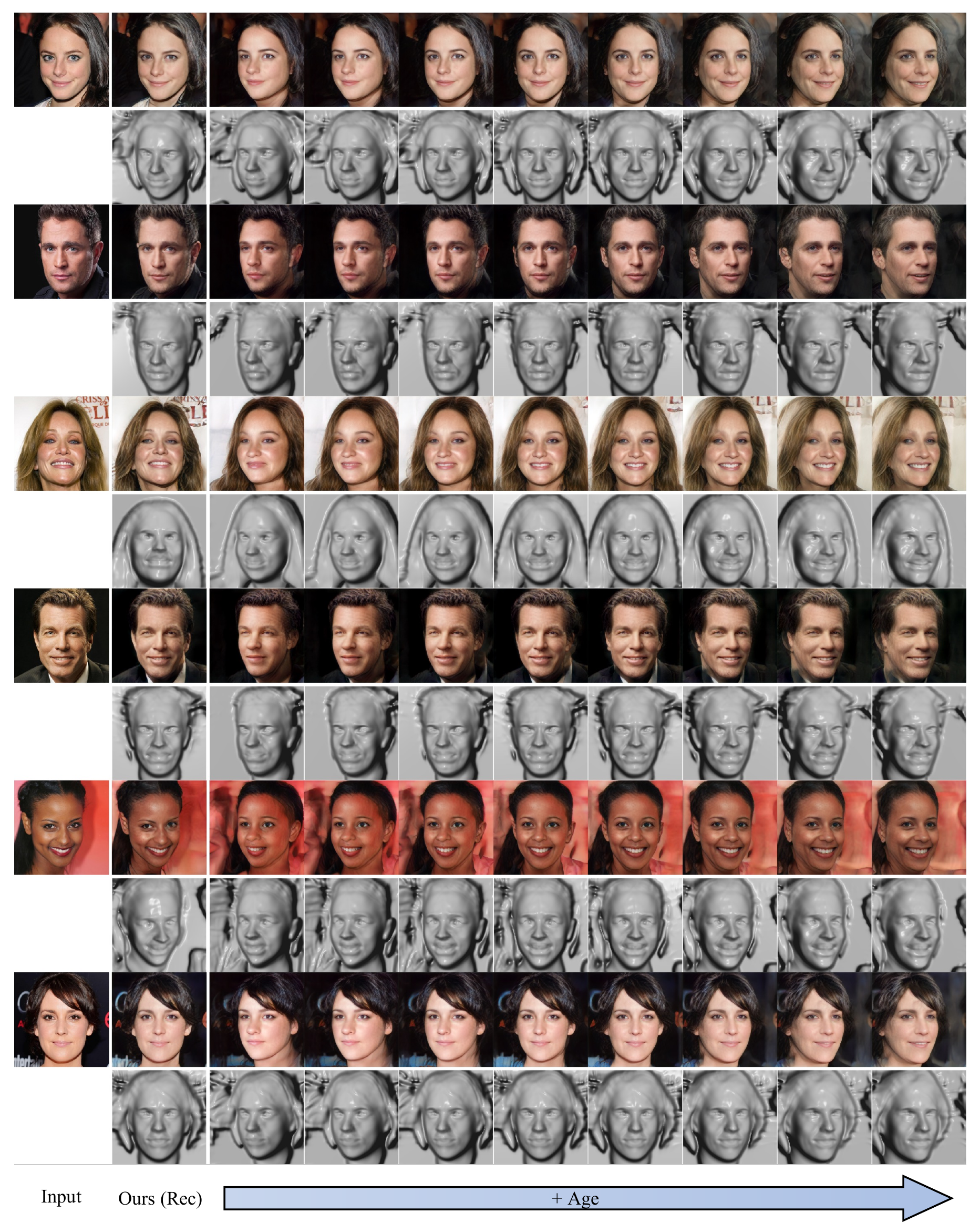}
\caption{Visual comparisons on face editing (Age).}
\label{fig:supp:editing:age}
\end{figure*}

\begin{figure*}[tp]
    \centering 
    \includegraphics[width=1\linewidth]{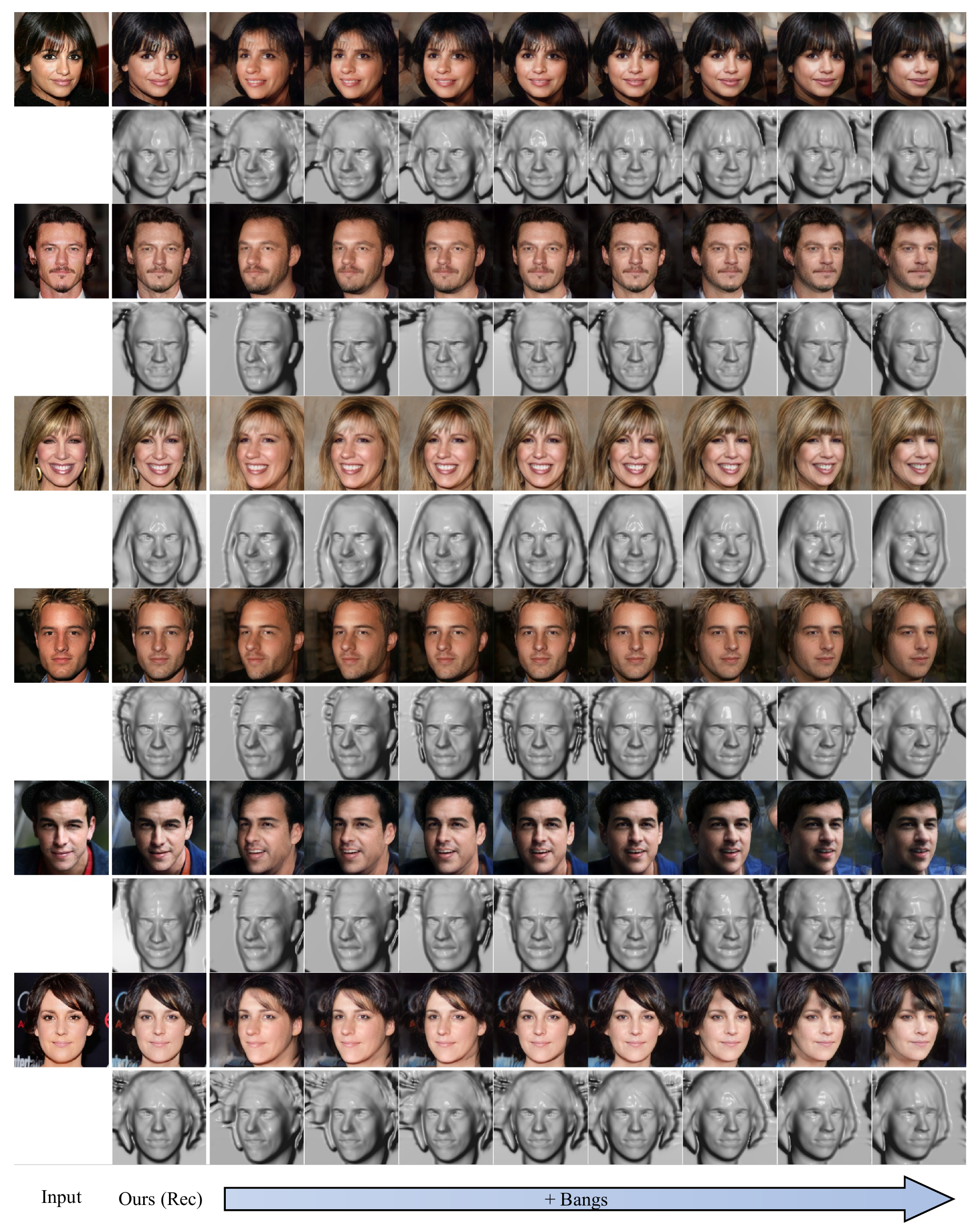}
\caption{Visual comparisons on face editing (Bangs).}
\label{fig:supp:editing:bangs}
\end{figure*}

\heading{More Editing Results}
We show more editing results on changing $4$ semantic attributes of our proposed method, 
namely smile (Fig.~\ref{fig:supp:editing:smile}), hair/beard (Fig.~\ref{fig:supp:editing:beard}), age (Fig.~\ref{fig:supp:editing:age}) and bangs (Fig.~\ref{fig:supp:editing:bangs}). 
Our method shows promising performance with shape-texture consistent editing. Note that since StyleSDF is still built on an MLP-based generator~\cite{Chan2021piGANPI} and InterfaceGAN~\cite{Shen2020InterFaceGANIT} is also not designed for 3D GANs,
the editing performance is hindered to some extent and cannot achieve comparable performance compared with 2D StyleGAN. 
However, we believe this limitation could be alleviated in the future by adopting better-designed 3D GAN architecture, \eg, tri-plane~\cite{Chan2021EG3D} and vision transformer~\cite{dosovitskiy2020vit}. Our results unleash the potential of this field and show that 3D consistency and high-fidelity reconstruction with high-quality editing are also achievable in recently developed 3D GAN. We hope our method could inspire later work in this field.

\begin{figure*}[tp]
    \centering 
    \includegraphics[width=0.99\linewidth]{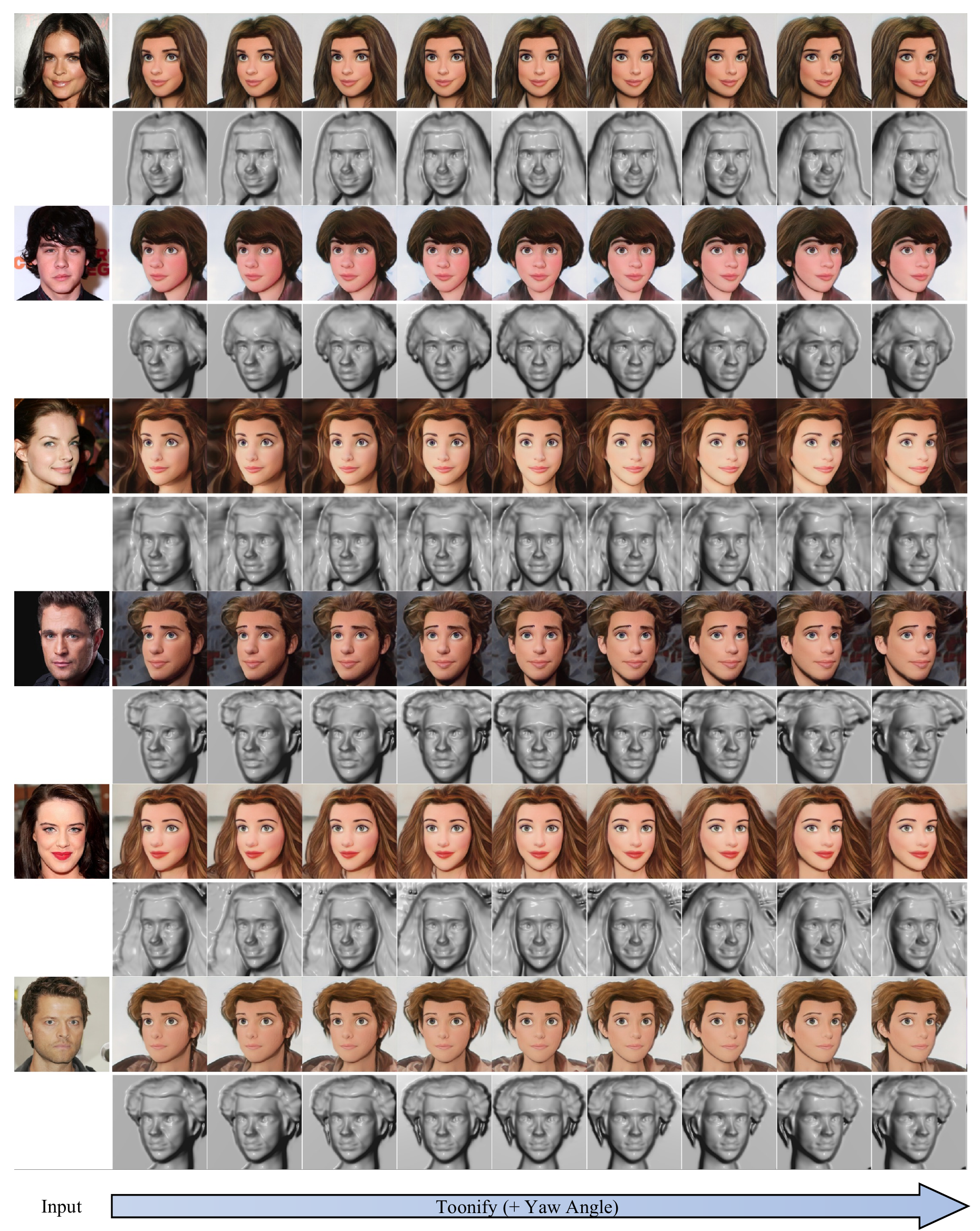}
\caption{Toonify results on faces.}
\label{fig:supp:toonify}
\end{figure*}

\heading{More Toonify Results}
We show 3D toonify-stylized results over real-world faces using our proposed method in Fig.~\ref{fig:supp:toonify}.
Following~\cite{pinkney2020resolution}, we finetune the pre-trained generator $\generator$ for $400$ iterations with $317$ cartoon face images and use our pre-trained encoder $\encoder$ for inference. 
Visually inspected, the toonified results holds the cartoon style and also preserve identity of the input image, which demonstrates the potential of applying our method over downstream tasks.

\clearpage
\clearpage

{\small
\bibliographystyle{ieee_fullname}
\bibliography{bibs/supplementary.bib}
}